\documentclass[review]{elsarticle}
\usepackage{diagbox,multirow,booktabs,makecell,bm,stmaryrd,makecell,pifont,colortbl}
\usepackage{subcaption,array,graphicx,rotating,amsmath,mathtools,relsize,enumitem}
\usepackage{breqn,amssymb,upgreek}
\usepackage[dvipsnames]{xcolor}
\newcolumntype{?}{!{\vrule width 1.2pt}}
\usepackage{lineno,hyperref}
\modulolinenumbers[5]
\newcommand{\vect}[1]{\boldsymbol{\mathbf{#1}}}

\usepackage{algcompatible}
\usepackage{algorithm}
\usepackage{algorithmicx}
\usepackage[noend]{algpseudocode}
\algnewcommand{\LineComment}[1]{\State\unskip\the\therules \(\triangleright\) #1}

\DeclareFontFamily{U}{jkpmia}{}
\DeclareFontShape{U}{jkpmia}{m}{it}{<->s*jkpmia}{}
\DeclareFontShape{U}{jkpmia}{bx}{it}{<->s*jkpbmia}{}
\DeclareMathAlphabet{\mathfrak}{U}{jkpmia}{m}{it}
\SetMathAlphabet{\mathfrak}{bold}{U}{jkpmia}{bx}{it}

\usepackage{array}
\newcommand{\PreserveBackslash}[1]{\let\temp=\\#1\let\\=\temp}
\newcolumntype{C}[1]{>{\PreserveBackslash\centering}p{#1}}
\newcolumntype{R}[1]{>{\PreserveBackslash\raggedleft}p{#1}}
\newcolumntype{L}[1]{>{\PreserveBackslash\raggedright}p{#1}}

\DeclareMathOperator*{\argmin}{arg\,min}
\DeclareMathOperator*{\argmax}{arg\,max}

\newcommand*\colourcheck[1]{%
  \expandafter\newcommand\csname #1check\endcsname{\textcolor{#1}{\ding{52}}}%
}
\newcommand*\colourcross[1]{%
  \expandafter\newcommand\csname #1cross\endcsname{\textcolor{#1}{\ding{55}}}%
}
\colourcheck{blue}
\colourcheck{OliveGreen}
\colourcheck{red}
\colourcross{blue}
\colourcross{green}
\colourcross{red}

\makeatletter
\def\therule{\makebox[\algorithmicindent][l]{\hspace*{.5em}\vrule height .75\baselineskip depth .25\baselineskip}}%

\newtoks\therules
\therules={}
\def\appendto#1#2{\expandafter#1\expandafter{\the#1#2}}
\def\gobblefirst#1{
  #1\expandafter\expandafter\expandafter{\expandafter\@gobble\the#1}}%
\def\LState{\State\unskip\the\therules}
\def\pushindent{\appendto\therules\therule}%
\def\popindent{\gobblefirst\therules}%
\def\printindent{\unskip\the\therules}%
\def\printandpush{\printindent\pushindent}%
\def\popandprint{\popindent\printindent}%

\algdef{SE}[WHILE]{While}{EndWhile}[1]
  {\printandpush\algorithmicwhile\ #1\ \algorithmicdo}
  {\popandprint\algorithmicend\ \algorithmicwhile}%
\algdef{SE}[FOR]{For}{EndFor}[1]
  {\printandpush\algorithmicfor\ #1\ \algorithmicdo}
  {\popandprint\algorithmicend\ \algorithmicfor}%
\algdef{S}[FOR]{ForAll}[1]
  {\printindent\algorithmicforall\ #1\ \algorithmicdo}%
\algdef{SE}[LOOP]{Loop}{EndLoop}
  {\printandpush\algorithmicloop}
  {\popandprint\algorithmicend\ \algorithmicloop}%
 \algdef{SE}[REPEAT]{Repeat}{Until}[1]
  {\printandpush\algorithmicrepeat\ #1}
  {\popandprint\algorithmicend}%
\algdef{SE}[IF]{If}{EndIf}[1]
  {\printandpush\algorithmicif\ #1\ \algorithmicthen}
  {\popandprint\algorithmicend\ \algorithmicif}%
\algdef{C}[IF]{IF}{ElsIf}[1]
  {\popandprint\pushindent\algorithmicelse\ \algorithmicif\ #1\ \algorithmicthen}%
\algdef{Ce}[ELSE]{IF}{Else}{EndIf}
  {\popandprint\pushindent\algorithmicelse}%
\algdef{SE}[PROCEDURE]{Procedure}{EndProcedure}[2]
   {\printandpush\algorithmicprocedure\ \textproc{#1}\ifthenelse{\equal{#2}{}}{}{(#2)}}%
   {\popandprint\algorithmicend\ \algorithmicprocedure}%
\algdef{SE}[FUNCTION]{Function}{EndFunction}[2]
   {\printandpush\algorithmicfunction\ \textproc{#1}\ifthenelse{\equal{#2}{}}{}{(#2)}}%
   {\popandprint\algorithmicend\ \algorithmicfunction}%

\journal{Journal of \LaTeX\ Templates}

\biboptions{sort&compress}
\begin{document}

\begin{frontmatter}

\title{Self-Supervised Leaf Segmentation under Complex Lighting Conditions}





\author[DeakinIT]{Xufeng Lin\corref{mycorrespondingauthor}}
\cortext[mycorrespondingauthor]{Corresponding author}
\ead{xufeng.lin@deakin.edu.au}

\author[DeakinIT]{Chang-Tsun Li}
\ead{changtsun.li@deakin.edu.au}

\author[DeakinEngineering]{Scott Adams}
\ead{scott.adams@deakin.edu.au}

\author[DeakinEngineering]{Abbas Kouzani}
\ead{abbas.kouzani@deakin.edu.au}

\author[Lancaster]{Richard Jiang}
\ead{r.jiang2@lancaster.ac.uk}

\author[Warwick]{Ligang He}
\ead{ligang.he@warwick.ac.uk}

\author[SCUT]{Yongjian Hu}
\ead{eeyjhu@scut.edu.cn}

\author[DeakinEngineering]{Michael Vernon}
\ead{m.vernon@deakin.edu.au}
\author[DeakinEngineering]{Egan Doeven}
\ead{egan.doeven@deakin.edu.au}
\author[DeakinLife]{Lawrence Webb}
\ead{lawrence.webb@deakin.edu.au}

\author[Medigrowth]{Todd Mcclellan}
\ead{todd@medigrowth.com.au}
\author[Medigrowth]{Adam Guskich}
\ead{adam@medigrowth.com.au}


\address[DeakinIT]{School of Information Technology, Deakin University, Waurn Ponds, Australia}
\address[DeakinEngineering]{School of Engineering, Deakin University, Waurn Ponds, Australia}
\address[DeakinLife]{School of Life and Environmental Sciences, Deakin University, Waurn Ponds, Australia}
\address[Warwick]{Department of Computer Science, The University of Warwick, Coventry, UK}
\address[Lancaster]{School of Computing and Communications, Lancaster University, Lancaster, UK}
\address[SCUT]{School of Electronic and Information Engineering, South China University of Technology, Guangzhou, China}
\address[Medigrowth]{Medigrowth Australia Pty Ltd, Waurn Ponds, Australia}

\begin{abstract}
As an essential prerequisite task in image-based plant phenotyping, leaf segmentation has garnered increasing attention in recent years. While self-supervised learning is emerging as an effective alternative to various computer vision tasks, its adaptation for image-based plant phenotyping remains rather unexplored. In this work, we present a self-supervised leaf segmentation framework consisting of a self-supervised semantic segmentation model, a color-based leaf segmentation algorithm, and a self-supervised color correction model. The self-supervised semantic segmentation model groups the semantically similar pixels by iteratively referring to the self-contained information, allowing the pixels of the same semantic object to be jointly considered by the color-based leaf segmentation algorithm for identifying the leaf regions. Additionally, we propose to use a self-supervised color correction model for images taken under complex illumination conditions. Experimental results on datasets of different plant species demonstrate the potential of the proposed self-supervised framework in achieving effective and generalizable leaf segmentation\footnote{The developed code and datasets will be made publicly available on \href{https://github.com/lxfhfut/Self-Supervised-Leaf-Segmentation}{https://github.com/lxfhfut/Self-Supervised-Leaf-Segmentation}}. 
\end{abstract}


\end{frontmatter}


\section{Introduction}
\label{sec:intro}
Plant phenotyping is the field of scientific inquiry concerned with the quantitative measurement of observable plant traits \cite{walter2015plant,walter2005dynamics,monforte2013genetic,arvidsson2011growth} developed from the dynamic interaction of the genotype with environmental conditions. It provides an important tool to understand the effects of environment on the cultivated plants, and enables a wide range of applications in plant breeding \cite{walter2015plant}, crop monitoring \cite{lee2010sensing,saiz2020smart}, disease prevention and control \cite{mutka2015image,mutka2016quantitative}, etc. While traditional plant phenotyping relies on labor-intensive and error-prone manual measurements, the advances in digital imaging and computer vision techniques have allowed for quantifying plant traits from images in a non-invasive and automatic manner. In achieving the goal of image-based plant phenotyping, automated segmentation of plant leaves is the fundamental prerequisite for measuring more complex phenotypic traits. It is often performed at two granular levels: category-level and instance-level. The former is concerned with segmenting the pixels belonging to the `leaf' category from background, while the latter moves a step further and separates individual leaves from each other. Instance-level leaf segmentation allows for fine-grained measurement of individual leaf area, leaf count and leaf growth rate, which could be beneficial for responsive plant growth monitoring and growth regulation \cite{ojolo2018regulation}. However, the variability in leaf shape and appearance, constant self-occlusion and varying imaging conditions often render instance-level leaf segmentation an extremely challenging problem even in controlled environments. In comparison, category-level leaf segmentation is relatively easier and provides a good approximation of plant size, thus making it a more feasible and practical means for many application scenarios, such as plant growth monitoring \cite{li2020measuring} and yield prediction \cite{van2020crop}. 


Facilitated by the Leaf Segmentation Challenge (LSC) of the Computer Vision Problems in Plant Phenotyping (CVPPP) workshop, significant advances have been achieved for both category-level and instance-level leaf segmentation. Focusing on rosette plants \cite{CVPPP_LSC}, the LSC challenge has been instrumental in advancing the research in leaf segmentation and beyond within the application domain of image-based plant phenotyping. While earlier works rely on hand-crafted image features \cite{3dHistogram,collation,pape2015utilizing}, state-of-the-art methods \cite{deepcoloring,synthetic,UPGen} are mainly based on supervised training of deep convolutional neural networks (CNNs), particularly U-net \cite{ronneberger2015u} and Mask-RCNN \cite{he2017mask}, which have demonstrated superior performance in segmenting common objects, e.g., person and car. However, despite the substantial progress made over the years, there are still many challenges hindering the wide applicability of existing techniques in practical deployment: 
\begin{itemize}[leftmargin=*]
\item Firstly, the training of deep models requires a large amount of annotated data, but obtaining pixel-wise annotation for segmentation could be a highly labor-intensive, time-consuming and error-prone process. For the image-based plant phenotyping problem at hand, annotated data must contain sufficient examples of different mutations, genotypes and environmental conditions covering different growth stages, which makes the problem thornier than expected. 
\item Secondly, deep learning models trained on datasets of specific plant species usually do not generalize well to other unseen species. This problem is particularly acute for plant leaf segmentation because different plant species vary dramatically in leaf appearances \cite{augmentation,UPGen}. It is often required to re-train a model from scratch or fine-tune a pre-trained model on annotated datasets of unseen plant species to achieve satisfactory performance across plant species. 
\item Lastly, dramatic changes in the background and appearance of plant leaves caused by varying lighting conditions adds another dimension to the challenges faced by image-based plant phenotyping. 
Along with other factors such as weather conditions or different times of the day, this will result in complex lighting conditions for plant image acquisition, and pose a huge challenge for leaf segmentation. Surprisingly, to the best of our knowledge, there has been limited research on investigating the effectiveness of leaf segmentation under different lighting conditions. Given that the use of artificial light has become very common in greenhouse cultivation to supplement natural sunlight, this is a significant omission as it precludes the use of many existing algorithms on plants grown under artificial lighting conditions. 
\end{itemize}

To mitigate the aforementioned problems, there have been many attempts to generate new synthetic data samples by 3D plant modelling \cite{ubbens2018use}, Generative Adversary Networks (GANs) \cite{valerio2017arigan,zhu2018data}, domain randomization \cite{synthetic,augmentation,UPGen}, etc. However, it is difficult and often impossible to accurately simulate different plant characteristics, various environmental conditions, and complex interplay between genetic and environmental factors, which inevitably creates a gap between the real and synthetic data. In this work, we propose to surmount the above challenges by developing a self-supervised learning framework for leaf segmentation under complex lighting conditions without using any annotated data. Specifically, we make the following contributions:
\begin{enumerate}[leftmargin=*]
    \item We propose a novel self-supervised semantic segmentation model. It integrates the feature extraction power of Convolutional Neural Networks (CNNs) with the structured modeling capabilities of fully connected Conditional Random Fields (CRFs). It allows the pixels of the same semantic object to be jointly processed, thus significantly reducing the impact of complex backgrounds and variations within the leaf and non-leaf regions.
    \item We propose a color-based leaf segmentation algorithm. It models the ``greenness'' of semantic objects in an image with the multivariate normal distribution in the HSV color space, and identifies the regions with admissible \emph{absolute} and \emph{relative} greenness as leaf regions. 
    \item We propose a self-supervised color correction model to rectify the ``distorted'' color in an image caused by the use of artificial grow lights. In so doing, the color-corrected images can be segmented in the same way as for the images taken under ``natural'' daylight conditions.
    \item We publish a dataset of top-view cannabis images captured in a greenhouse equipped with grow lights to facilitate the research in the area of image-based plant phenotyping.
\end{enumerate}

The remainder of this manuscript is organized as follows. We first review the literature relevant to the proposed method in Section 2. The details of the proposed self-supervised leaf segmentation framework are presented in Section 3, followed by comprehensive experimental results in Section 4. The concluding remarks along with a discussion of future works are given in Section 5.

\section{Related Works}
\label{sec:related_work}
\noindent\textbf{Unsupervised leaf segmentation}. Unsupervised image segmentation aims to partition an image into groups of perceptually or semantically similar pixels without resorting to the ground-truth annotations. In the specific case of leaf segmentation, traditional unsupervised clustering algorithms, e.g., expectation maximization (EM) algorithm \cite{leafsnap}, K-means \cite{zhang2019plant}, and fuzzy clustering \cite{bai2017fuzzy}, based on color \cite{leafsnap,cerutti2013understanding,zhang2019plant}, shape \cite{cerutti2013understanding}, texture \cite{zou2019broccoli} features have been widely adopted to distinguish the ``leaf'' pixels from the background. These methods are usually employed in conjunction with superpixel algorithms to enhance the spatial consistency and boundary adherence of the segmentation result. As can be expected, these methods inevitably inherit the shortcoming of being sensitive to parameters and outliers from the adopted traditional clustering algorithms. Consequently, tedious parameter tuning and ad-hoc post-processing are usually required to obtain satisfactory segmentation results on specific datasets. Moreover, the image features, particularly the shape and texture features, that these unsupervised leaf segmentation algorithms rely on are specially designed for a specific plant species and do not generalize well across a variety of plant species. One may argue that the color feature is generalizable over different plant species because the leaves of most plants are green due to the presence of chlorophyll. Indeed, the color feature is arguably the most widely used feature in plant-related image analysis. However, the color-based leaf segmentation can be significantly influenced by the lighting conditions and the green-looking objects present in the background, e.g., mosses and weeds. 

\noindent\textbf{Supervised leaf segmentation}. Supervised leaf segmentation aims to segment leaf pixels in an image with a model trained on annotated image datasets. While some early works \cite{zheng2010segmentation,3dHistogram} attempted to accomplish this task by learning the distributions of leaf and non-leaf pixels in the color space, the past few years have seen intensive use of methods based on deep neural networks in various computer vision tasks, with no exception for image-based plant phenotyping. Deep neural network architectures, such as U-Net \cite{Unet} and Mask R-CNN \cite{he2017mask}, have been successfully used for category-level \cite{baweja2018stalknet,fawakherji2019crop,yang2020semantic} or instance-level \cite{synthetic,UPGen,augmentation,yang2020leaf} leaf segmentation. To harness the full potential of deep neural networks, it is essential to train the networks on large-scale high-quality annotated datasets. However, the expense of the specialized facilities and equipment for growing and monitoring individual plants, have substantially hindered the collection and annotation of large representative datasets required in training deep learning models for image-based plant phenotyping. To mitigate the data scarcity issue, Ward \emph{et al.} \cite{synthetic,UPGen} employed domain randomization to generate synthetic arabidopsis leaf images. With a pool of ``inspiration'' leaves with leaf geometries, leaf textures, and backgrounds collected from existing annotated leaf image datasets, a synthetic plant is generated by randomizing the background and various plant parameters, e.g., leaf shapes and textures, sampled from the pool of inspiration leaves. A similar idea was proposed by Kuznichov \emph{et al.} \cite{augmentation}, where a synthetic image is generated by applying geometric transformations with random parameters to individual leaves segmented from real leaf images and pasting them in random (\emph{na\"ive collage}) or logical and structured (\emph{structured collage}) locations over a background image randomly selected from the CVPPP LSC dataset \cite{CVPPP_LSC}. Some other works \cite{valerio2017arigan,zhu2018data,barth2020optimising} resort to generative adversarial networks (GANs) \cite{goodfellow2014generative} to generate synthetic plant images. While effective in generating realistic synthetic plant images, all the above-mentioned methods are still highly reliant on large amounts of annotated images.

\noindent\textbf{Self-supervised leaf segmentation}. Self-supervised learning aims to automatically generate some kind of supervisory signal, e.g., pseudo labels, from unlabeled data to solve tasks that are typically targeted by supervised learning. As the supervisory signal is automatically generated from the data itself or its transformed versions, self-supervised learning does not rely on human labeled data and thus can be considered as a subset of unsupervised learning. To differentiate it from traditionally unsupervised learning (e.g., K-Means and fuzzy clustering), in this work, we use the term ``self-supervised learning'' to refer to the techniques that explicitly and automatically generate supervisory signals for typical supervised learning tasks such as classification and regression. Through solving pretext tasks \cite{larsson2017colorization,pathak2016context}, self-supervised learning has been widely used to pre-train deep neural networks for learning visual representations that can be transferred to downstream tasks, e.g., image classification \cite{van2020scan}, object detection \cite{misra2020self}, and semantic segmentation \cite{van2020scan}. There has also been a recent emergence of self-supervised methods that directly output class labels for image clustering and semantic segmentation without the use of a pretext task. For instance, Ji \emph{et al.} \cite{ji2019invariant} trained a deep neural network by maximizing the mutual information between the network outputs of an image and its augmented versions to predict the image-level and pixel-level semantic labels for image clustering and image segmentation, respectively. In a similar vein, some works attempted to learn pixel-level representations for semantic segmentation from different views \cite{ouali2020autoregressive} or object mask proposals \cite{van2021unsupervised} of the input image. Another line of self-supervised semantic segmentation \cite{kanezaki2018unsupervised,kim2020unsupervised} constructs supervisory signal by grouping spatially adjacent pixels with similar features and iteratively updating the network parameters until semantic label assignment converges. However, despite the growing popularity in machine learning community, self-supervised learning is still fairly unexplored in image-based plant phenotyping and, more broadly, in agricultural technology, with a few exceptions that leverage self-supervised methods for pre-training agricultural image classification model, e.g., \cite{guldenring2021self}.

\begin{figure}[]
    \begin{subfigure}{0.35\textwidth}
            \includegraphics[width=0.95\linewidth]{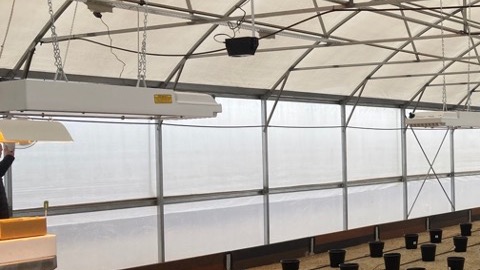}
            \subcaption{Greenhouse setup}
            \label{subfig:greenhouse}
    \end{subfigure}%
    \begin{subfigure}{0.3\textwidth}
            \includegraphics[width=0.95\linewidth]{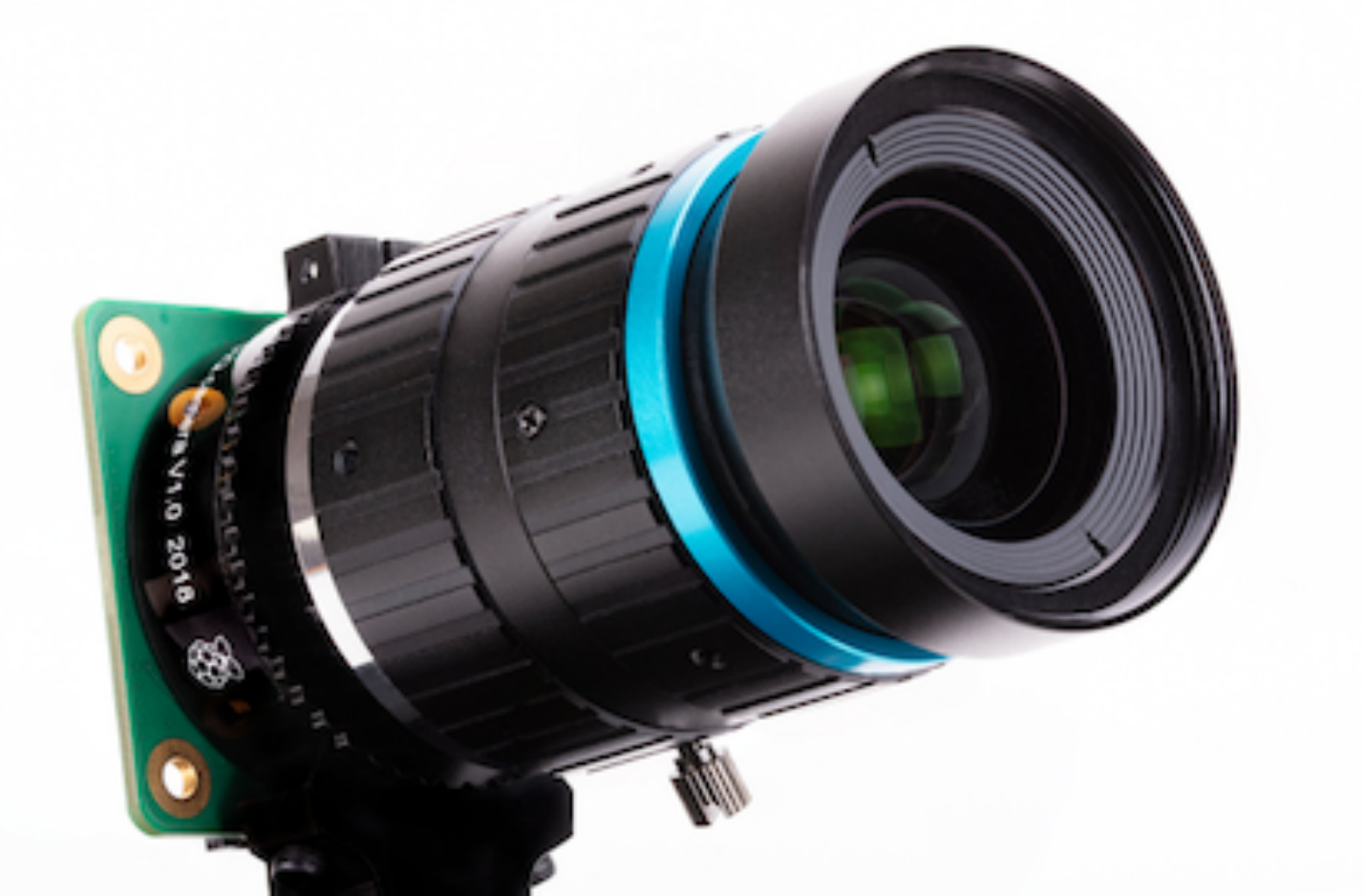}
            \subcaption{Camera}
            \label{subfig:camera_sensor}
    \end{subfigure}%
    \begin{subfigure}{0.35\textwidth}
            \includegraphics[width=0.95\linewidth]{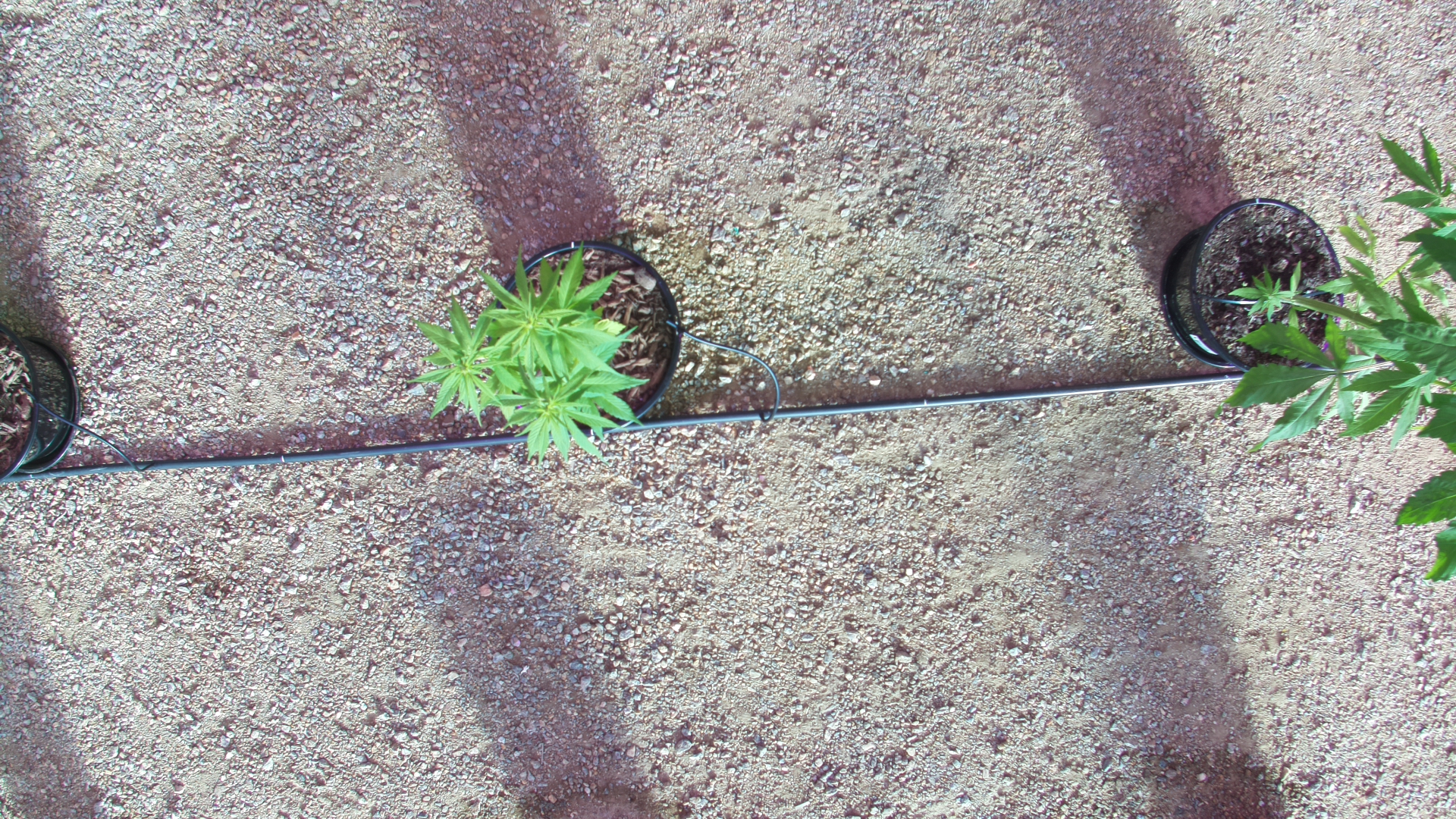}
            \subcaption{Sample top-view image}
            \label{subfig:camera}
    \end{subfigure}
\caption{Image acquisition setup for our Cannabis dataset.}
\label{fig:image_acquisition_setup}
\end{figure}

\section{Methodology}
\label{sec:method}
\begin{figure}[!t]
\centering
\includegraphics[width=0.99\textwidth]{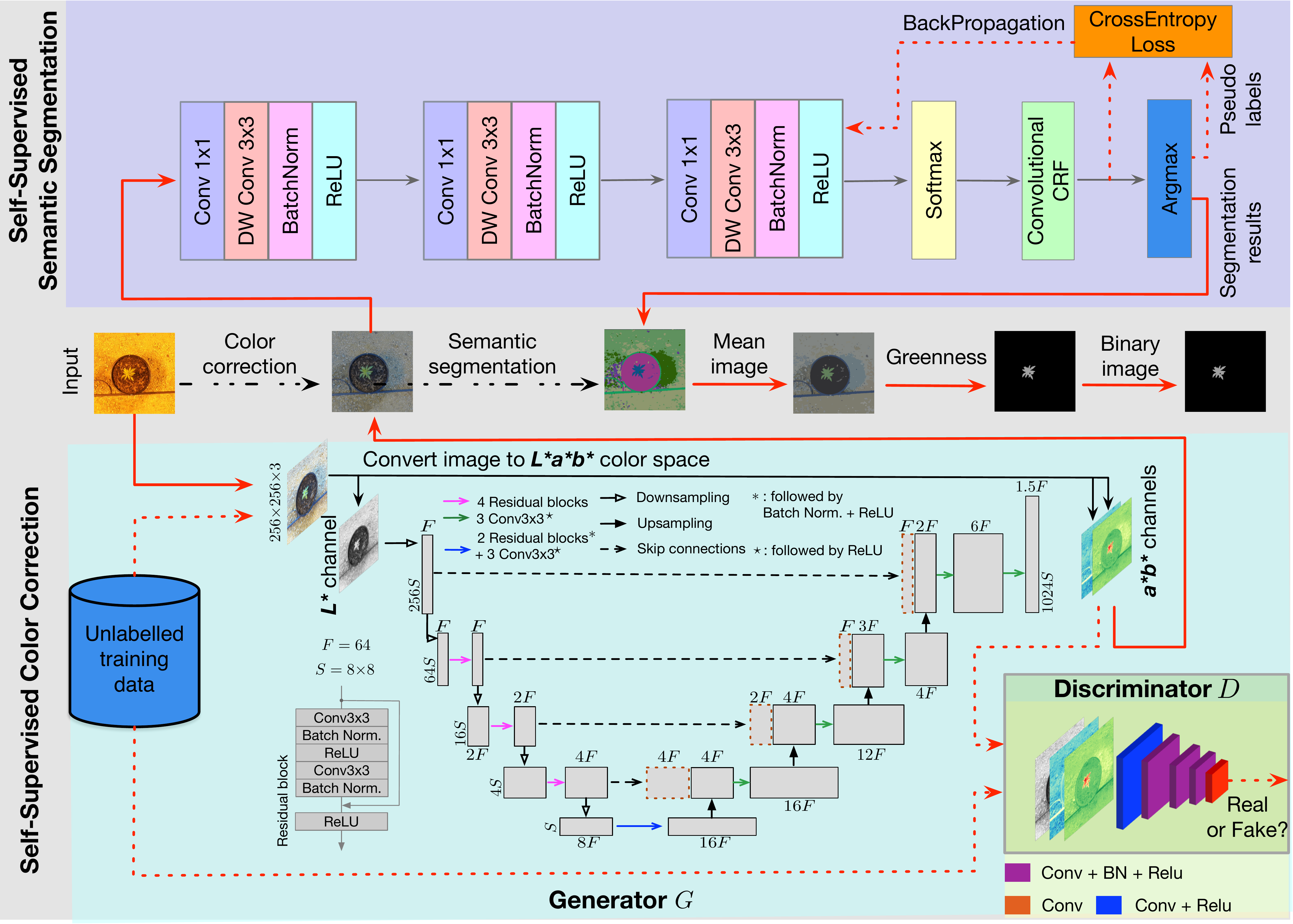}
\caption{Overview of the proposed self-supervised leaf segmentation framework. It mainly consists of three components: self-supervised color correction, self-supervised semantic segmentation, and color-based leaf segmentation. The data flows during the training and testing phases are shown with red dashed and red solid lines, respectively. An image input is first passed through the trained color correction model to rectify the potential `unnatural' color in the image. The color-corrected image is then input to the self-supervised semantic segmentation model to group the pixels of semantically similar objects, which will be jointly considered to identify the green leaf objects with our color-based leaf segmentation model.}
\label{fig:flowchart}
\end{figure}

We first describe the unique opportunity we had to design and deploy a cannabis growth monitoring system in a real-world greenhouse environment. As shown in Fig. \ref{fig:image_acquisition_setup}, cameras are mounted directly above the pots to collect top-view images at an hourly interval over the course of the whole growth cycle. Depending on the growth stage of the cannabis, the growing environment can be customized by controlling the temperature, humidity and lighting conditions to optimize the cannabis yield. We would like to particularly stress that the lighting schedule (i.e., exposure duration) and quality (i.e., light spectrum) can directly impact the transition between growth stages and ultimately affect the yield of the plant. Thus, a high-quality artificial lighting environment is critical for the effective indoor cultivation of cannabis. In such a scenario, measuring the leaf area index of plant canopy, which can be approached via category-level leaf segmentation, is practically more feasible than counting the number of leaves for growth monitoring due to the heavy leaf occlusions that often occur at the later growth stages of the cannabis. The key challenges for leaf segmentation in this typical scenario are twofold. First, while data can be collected round the clock automatically, annotating the collected data for training leaf segmentation algorithms is labor-intensive and error-prone. Second, the use of artificial grow lights poses a great challenge for many segmentation algorithms as it dramatically changes the appearance of the plant in the image. In this work, we propose a self-supervised leaf segmentation framework that provides a promising way towards effective and generalizable leaf segmentation under complex lighting conditions without the need for annotated data. Fig. \ref{fig:flowchart} shows the overview of our proposed framework, which mainly consists of three components: self-supervised color correction, self-supervised semantic segmentation, and color-based leaf segmentation. In what follows, we will delve into the details of the self-supervised semantic segmentation and color-based leaf segmentation for the images acquired under ``natural'' or ``normal'' lighting conditions. We then introduce the self-supervised color correction model for correcting the color of the images taken under ``unnatural'' artificial lights so that the color-corrected images can be segmented in the same way as for ``natural'' images.

\subsection{Pseudo Label Generation for Self-Supervised Semantic Segmentation}
\label{subsec:pseudo_label_generation}
At the core of mainstream self-supervised semantic segmentation approaches is the generation of supervisory signals, typically in the form of ``pseudo labels'' for samples of the same or different classes, by leveraging human prior knowledge lying in the data. Images and their augmentations are used for generating positive samples of the same class, while all other images are considered as negative samples. With the positive and negative samples generated from a large number of images, a convolutional neural network (CNN) can be trained to extract pixel-level embeddings or representations for predicting semantic labels. Taking a detour from this prevalent practice, we approach the self-supervision problem by letting the neural network itself determine whether two pixels (actually two local patches due to the spatial locality enforced by the convolutional operation) of the same image belong to the same class or not. The underlying assumption is that \emph{semantically similar pixels should be mapped by an appropriately parameterized embedding learning network into representation embeddings that are close to each other in the embedding space, and therefore are more likely to be assigned with the same semantic label.} 

Concretely, for a pixel-level embedding learning function $\Phi_{\theta}: \mathcal{X} \rightarrow \mathcal{Z}$ parameterized by a neural network with weights $\theta$, it is expected to map two similar image pixels $\mathbf{x},\mathbf{x^\prime}\in\mathcal{X}$ to two similar representation embeddings $\mathbf{z},\mathbf{z^\prime}\in\mathcal{Z}$. A shallow network with 2 or 3 convolutional layers is sufficient to extract discriminative features for our task of leaf segmentation, as shown in Fig. \ref{fig:flowchart}. Following \cite{howard2017mobilenets}, we replace the traditional $3\times3$ convolution with a $1\times1$ pointwise convolution and a $3\times3$ depthwise convolution, which reduces the number of parameters of the network and speeds up the embedding learning process. Similar to \cite{ji2019invariant}, we terminate the output of convolutional layers with a softmax layer, which allows us to model the uncertainty of the label assignment at the pixel $i$ with a discrete probability distribution over $K$ semantic labels, i.e., $\mathbf{z}_i=\Phi_{\theta}(\mathbf{x}_i)\in[0,1]^K$. As such, the supervisory signals can be generated by exploiting the information readily available in the discrete probability distributions. A straightforward way to obtain the ``pseudo labels'', as proposed in \cite{kanezaki2018unsupervised,kim2020unsupervised}, is to perform the \emph{argmax} classification by assigning the index with the highest probability to each pixel, i.e., $l_i=\{j^{\star}|\mathbf{z}_{i,j^{\star}}{=}\max_{j} \mathbf{z}_{i,j}\}$, where $l_i$ is the pseudo label for the pixel $i$ and $\mathbf{z}_{i,j}$ is the $j$th element of $\mathbf{z}_i$. These pixel-level pseudo labels are, in turn, used in a supervised fashion to update the network parameters via backpropagation. This procedure is repeated iteratively until convergence or the maximum number of iterations is reached. 

\subsection{Challenges of Pseudo Label Refinement}
\label{subsec:challenges_label_refinement}
As in supervised learning, the quality of the pseudo labels also has a significant impact on the performance of self-supervised learning approaches. However, the aforementioned way of pseudo label generation, in its primitive form, is prone to noisy labels possibly due to the intrinsic properties of convolutional neural networks, e.g., the sensitivity to small perturbations in the image \cite{szegedy2013intriguing} and the tendency to output blurry object boundaries \cite{chen2014semantic}. Consequently, semantically similar pixels may be assigned with different labels, while pixels of different semantic objects are likely to be assigned with the same label. These problems are more prominent in the earlier stage of the iterative procedure when the network weights are primarily random values. 

Two approaches have been pursued to impose additional constraints for the refinement of pseudo label assignment. The first approach \cite{kanezaki2018unsupervised} is to apply superpixel segmentation (e.g., SLIC \cite{achanta2012slic}) beforehand and force the pixels in the same superpixel to have the same pseudo label. The second approach is to employ a spatial continuity loss \cite{kim2020unsupervised} to encourage consistent pseudo label assignment for adjacent pixels. The drawback of the first approach is that superpixel segmentation itself is an ill-posed problem and the errors in superpixel segmentation, which often occur in object boundaries, may lead to inaccurate and misleading pseudo label assignment. Moreover, as the superpixel segmentation is only performed once prior to the iterative update of the network parameters, it does not allow the local neighborhood information conveyed by superpixels to be updated in a dynamic way. Inevitably, little useful information can be provided for pseudo label refinement once the spatial consistency enforced by superpixels has been fulfilled. For the second approach, the spatial continuity loss encourages spatial consistency of label assignment by enforcing the extracted representation embeddings of adjacent pixels to be close to each other. Not only does it neglect the long-range dependencies between pixels, it also disregards the fact that adjacent pixels may belong to different semantic objects. Such a boundary-unaware label propagation may provide conflicting information for label refinement and often results in subtle segments within the same semantic object, which can be clearly observed in the visualization results presented in \cite{kim2020unsupervised}.

\subsection{Fully-Connected CRFs for Structured Pseudo Label Refinement}
\label{subsec:crf_refinement}
In response to the above-mentioned limitations, we propose to integrate the fully connected conditional random field (CRF) \cite{krahenbuhl2011efficient} into the iterative label assignment procedure of our self-supervised semantic segmentation model. Given an image $\mathbf{X}$ consisting of $N$ pixels, we model its segmentation as a random field defined over a set of variables $\bm{L}=\{l_1, l_2, ..., l_N\}$, where $l_i$ represents the label assigned to the pixel $i$ and can take any value from a set of $K$ semantic labels $\mathcal{L}=\{1, 2, ... , K\}$. A conditional random field $(\mathbf{X}, \bm{L})$ can be characterized by a Gibbs distribution in the form of $P(\bm{L}{=}\bm{l}|\mathbf{X})=\frac{1}{Z(\mathbf{X})}\exp(-E(\bm{l}|\mathbf{X}))$, where $E(\bm{l}|\mathbf{X})$ is the Gibbs energy of a labeling configuration $\bm{l}\in\mathcal{L}^N$ and $Z(\mathbf{X})$ is the partition function. For the fully connected CRF model in \cite{krahenbuhl2011efficient}, the Gibbs energy is given by
\begin{equation}
    E(\bm{l}|\mathbf{X}) = \sum_{i\leqslant N}{\psi_u(l_i|\mathbf{X})} + \eta\sum_{i\neq j\leqslant N}{\psi_p(l_i,l_j|\mathbf{X})},
\label{eqn:gibbs_energy}
\end{equation}
where the unary potential $\psi_u(l_i|\mathbf{X})$ measures the cost of assigning label $l_i$ to the pixel $i$ and the pairwise potential $\psi_p(l_i,l_j|\mathbf{X})$ measures the cost of assigning labels $l_i, l_j$ to pixels $i, j$ simultaneously. $\eta$ is a weighting factor adjusting the relative importance of the unary and pairwise potentials. 

For our pseudo label refinement, we set the unary potential as $\psi_u(l_i|\mathbf{X}){=}-\log z_{i, l_i}$, where $z_{i, l_i}$ is the probability of assigning label $l_i$ to pixel $i$ as output by the softmax layer of the embedding learning network $\Phi_{\theta}$. While for the pairwise potential $\psi_p(l_i,l_j|\mathbf{X})$, we adopt a Gaussian \emph{appearance} kernel \cite{krahenbuhl2011efficient}:
\begin{align}
\psi_p(l_i,l_j|\mathbf{X}) &=\mu(l_i,l_j) k(\mathbf{f}_i,\mathbf{f}_j)\nonumber \\ &=\mu(l_i,l_j)\underbrace{\exp{\biggl(-\frac{\|\mathbf{p}_i-\mathbf{p}_j\|_2^2}{2\sigma_{\alpha}^2}-\frac{\|\mathbf{x}_i-\mathbf{x}_j\|_2^2}{2\sigma_{\beta}^2}\biggr)}}_{\emph{appearance kernel}}.
\label{eqn:pairwise_potential}
\end{align}

The label compatibility function $\mu(l_i,l_j)$ imposes a penalty when different labels $l_i$ and $l_j$ are assigned to adjacent pixels. While it is possible to specify different penalties for different pairs of labels or make $\mu(l_i,l_j)$ as parameters that can be learned from the data as in \cite{krahenbuhl2011efficient}, it is unreasonable to do so in our self-supervised setting as the labels are randomly assigned for different images and there are no pre-determined semantic meanings for the labels. In other words, the leaf pixels in two different images may be represented by different labels. For this reason, we use the simple and most widely used Potts model given by $\mu(l_i,l_j)=\llbracket l_i\neq l_j \rrbracket$, where $\llbracket \rrbracket$ is the Iverson bracket.

The appearance kernel $k(\mathbf{f}_i,\mathbf{f}_j)$ in Eq. (\ref{eqn:pairwise_potential}) depends on both pixel locations $(\mathbf{p}_i, \mathbf{p}_j)$ and the corresponding color vectors $(\mathbf{x}_i, \mathbf{x}_j)$ in the RGB color space. Intuitively, it tends to assign the same label for adjacent pixels with similar color, with the ``scale'' of spatial distance and color proximity controlled by the parameters $\sigma_\alpha$ and $\sigma_\beta$. As each pair of pixels $i$ and $j$ will contribute to the pairwise potential, regardless of their distance from each other, the fully connected CRF model allows to exploit long-range pixel dependencies for pseudo label refinement. Note that in the original model in \cite{krahenbuhl2011efficient}, there is a \emph{smoothness} kernel intended for removing small isolated regions. We discard it in our method as it will give rise to the chance of merging small, even though visually distinct, into the background, which could be detrimental for separating small plant leaves (e.g., at the seeding stage) from the background.
\begin{algorithm}
\caption{Self-Supervised Semantic Segmentation}
\begin{algorithmic}[1]           
           \For {$t=1$ to $T$}
           \LState $\{\mathbf{q}_i\}_{i=1}^N, \{\mathbf{z}_i\}_{i=1}^N \leftarrow \{\Phi_\theta(\mathbf{x}_i)\}_{i=1}^N$ \Comment{Pixel-level softmax output}
           \Repeat{$m{=}5$ \textbf{times}}
           \LState $\{\tilde{{q}}_{i,l}\}_{i=1}^N \leftarrow \{\eta\sum_{j\neq i}k(\mathbf{f}_i, \mathbf{f}_j)\:{q}_{j,l}\}_{i=1}^N$ \Comment{Message passing}
           \LState $\{\hat{{q}}_{i,l}\}_{i=1}^N \leftarrow \{\sum_{l^{\prime}\in \mathcal{L}}\mu(l,l^{\prime})\:\tilde{{q}}_{i,l^{\prime}}\}_{i=1}^N$ \Comment{Compatibility transform}
           \LState $\{\check{{q}}_{i,l}\}_{i=1}^N \leftarrow \{\hat{{q}}_{i,l}+\log z_{i,l}\}_{i=1}^N$  \Comment{Adding unary potentials}
           \LState $\{\mathbf{q}_i\}_{i=1}^N \leftarrow \{\text{Softmax}(\check{\mathbf{q}}_{i})\}_{i=1}^N$  \Comment{Softmax normalization}
           \Until{}
           \LState $\{l_i\}_{i=1}^N \leftarrow \{\argmax_j q_{i,j}\}_{i=1}^N$  \Comment{Supervisory pseudo labels} 
           \LState $\mathfrak{L} \leftarrow \text{CrossEntropyLoss}(\{\mathbf{q}_i, l_i\}_{i=1}^N)$  \Comment{Cross-entropy loss}
           \LState $\Phi_\theta \leftarrow \text{Update}(\mathfrak{L},\Phi_\theta)$  \Comment{Network parameters update}
           \EndFor
           \LState \Return $\{l_i\}_{i=1}^N$
\end{algorithmic}
\label{alg:procedure1}
\end{algorithm}
Under such formalization, our pseudo label refinement for a given image $\mathbf{X}$ can be achieved by finding the most probable label assignment $\bm{l}^{\star}$ that gives the maximum a posteriori (MAP) labeling of the random field, i.e., $\bm{l}^{\star}=\argmax_{\bm{l}\in\mathcal{L}^N}{P(\bm{l}|\mathbf{X})}$, or equivalently, the lowest Gibbs energy $E(\bm{l}^{\star}|\mathbf{X})$. However, CRFs are notoriously hard to optimize \cite{wilson2003, LI2008} and the exact maximization of $P(\bm{l}|\mathbf{X})$ is intractable even for low-resolution images. To circumvent this issue, a mean-field algorithm was proposed in \cite{krahenbuhl2011efficient} for approximate MAP marginal inference of $P(\bm{l}|\mathbf{X})$. The basic idea is to approximate the distribution $P(\bm{l}|\mathbf{X})$ with a simpler distribution $Q(\bm{l}|\mathbf{X})$ that can be
expressed as a product of independent marginals. The details of the mean-field algorithm are summarized in Steps 2-8 of Algorithm \ref{alg:procedure1}, where the distribution $Q(\bm{l}|\mathbf{X})$ is initialized as the pixel-level softmax output $\{\mathbf{q}_i\}_{i=1}^N$ of the embedding learning network $\Phi_\theta$. 

The efficient implementation of the mean-field algorithm is important for our label refinement as it may be executed $T$ (a few hundreds) times until the neural network $\Phi_\theta$ has been trained to extract meaningful embeddings for semantic segmentation. Fortunately, it was shown in \cite{zheng2015conditional} that all steps of the mean-filed algorithm (i.e., the Steps 2-8 of Algorithm \ref{alg:procedure1}) can be efficiently implemented on GPUs. Of particular note is the message passing (the Step 4 of Algorithm \ref{alg:procedure1}), which can be implemented as a Gaussian filter with the coefficients calculated using the Gaussian appearance kernel in Eq. (\ref{eqn:pairwise_potential}). The fully connected CRF, in its original form, allows for modeling the dependency between any pair of pixels in an image, resulting in a Gaussian filter that potentially spans the whole image. As suggested in \cite{teichmann2018convolutional}, this issue can be circumvented by assuming that the label distributions of two pixels are conditionally independent if their Manhattan distance is greater than $k$. Such a conditional independence assumption allows to efficiently implement the message passing with a $k\times k$ convolutional filter while still retaining the capability and flexibility of modeling long-range pixel dependencies. As shown in Algorithm \ref{alg:procedure1}, the mean-filed algorithm is repeated $m=5$ times, as suggested in \cite{ronneberger2015u,teichmann2018convolutional}, to obtain the refined label assignment. Afterwards in the Steps 9-11 of Algorithm \ref{alg:procedure1}, the supervisory pseudo labels, generated by applying \emph{argmax} classification to the refined label assignment distribution of each pixel, are used to calculate the multi-class cross-entropy loss for updating the embedding network with backpropagation. The entire procedure is repeated $T$ times until the network $\Phi_\theta$ is capable of extracting meaningful embeddings.

\subsection{Color-Based Leaf Segmentation}
\label{subsec:greenness}
Most existing semantic segmentation algorithms, including many self-supervised methods, require large-scale image datasets for training the network to group pixels into a pre-defined set of semantic classes. In contrast, our proposed self-supervised semantic segmentation algorithm learns to assign the same label to the semantically similar pixels with the self-contained information in a single image. While this precludes the use of external data, the side effect is that additional efforts are required to distinguish the leaves from other objects. Prior works \cite{leafsnap,3dHistogram} have shown the potential of color-based features for leaf segmentation, albeit for images with homogeneous backgrounds or in the supervised setting. With the results output by our self-supervised semantic segmentation algorithm, we are allowed to jointly process similar pixels of the same semantic label and extract more reliable color information that is less susceptible to the cluttered backgrounds or the subtle changes in leaf pixels. Towards this end, we propose a leaf segmentation algorithm based on the ``greenness'' of the pixels. 

Specifically, we first replace each pixel color $\mathbf{x}_i$ with the mean color of the pixels with the same label in its \emph{connected} region, i.e., $\bar{\mathbf{x}}_i=\frac{1}{|\mathcal{X}_{l_i}|}\sum_{j\in\mathcal{X}_{l_i}}\mathbf{x}_j$, where $\mathcal{X}_{l_i}$ is the set of pixels with the label $l_i$ in the \emph{connected} region of pixel $i$. The use of \emph{connected} regions enforces the calculation of the mean color is performed locally, preventing two remotely located objects in the image from influencing each other. Next, we convert the image from RGB to the HSV color space (i.e., $\mathbf{v}_i=\text{rgb2hsl}(\bar{\mathbf{x}}_i)$) and measure the ``greenness'' of each pixel with the following multivariate normal distribution in the HSV color space:
\begin{equation}
g(\mathbf{v}_i)=\frac{1}{\sqrt{(2\pi)^3\;det(\mathbf{\Sigma})}}\exp\Bigl(-\frac{1}{2}(\mathbf{v}_i-\vect{\upmu})^T\mathbf{\Sigma}^{-1}(\mathbf{v}_i-\vect{\upmu})\Bigr),
\label{eqn:greenness}
\end{equation}
where $\vect{\upmu}\in\mathbb{R}^3$ and $\mathbf{\Sigma}$ are the user-specified mean color vector and diagonal covariance matrix. 
Finally, a binary leaf segmentation mask is generated by applying the following thresholding operation:
\begin{equation}
\hat{u}_i=
\begin{dcases}
1, \;\;\;\tilde{g}(\mathbf{v}_i) > \gamma_1 \;\;\text{AND}\;\; \breve{g}(\mathbf{v}_i) > \gamma_2\\
0, \;\;\;\text{otherwise},
\end{dcases}
\label{eqn:threshold}
\end{equation}
where
\begin{equation}
\begin{dcases}
\tilde{g}(\mathbf{v}_i)=\frac{g(\mathbf{v}_i)}{g(\vect{\upmu})} \in [0,1]\\
\breve{g}(\mathbf{v}_i)=\frac{g(\mathbf{v}_i)-\min_i {g(\mathbf{v}_i)}}{\max_i {g(\mathbf{v}_i)}-\min_i {g(\mathbf{v}_i)}} \in [0,1]
\end{dcases}
\label{eqn:absolute_relative_greeness}
\end{equation}
$\tilde{g}(\mathbf{v}_i)$ and $\breve{g}(\mathbf{v}_i)$ measure the \emph{absolute} and the \emph{relative} ``greenness'', respectively. Intuitively, the \emph{relative} greenness measures the relative degree of greenness of an object by comparing to the highest ($\max_i {g(\mathbf{v}_i)}$) and the lowest \emph{absolute} greenness ($\min_i {g(\mathbf{v}_i)}$) in the same image. It allows us to select only the most green objects in an image, which could be particularly useful for reducing false positives when non-leaf but green-looking objects (e.g., mosses) appear in the background.


\subsection{Self-Supervised Color Correction}
\label{subsec:color_correction}
The last building block of our leaf segmentation framework is the self-supervised color correction model. As shown in Fig. \ref{fig:flowchart}, our color correction model follows the GAN-based pixel2pixel image translation network architecture \cite{isola2017image}, with a generator responsible for generating color-corrected images and a discriminator responsible for discriminating ``real'' images taken under good lighting conditions and ``fake'' images generated by the generator. The generator $G$, mainly consisting of a series of Convolution-BatchNorm-ReLU modules and residual blocks, progressively downsamples the input image to obtain high-level features and then gradually upsamples the features to generate the target images. To compensate for the low-level information lost due to downsampling operations, skip connections are used as ``shortcuts'' to allow for the direct information flow between the downsampling and upsampling branches. The discriminator $D$ is a binary classifier and is constructed by simply stacking a few blocks of Convolution-BatchNorm-ReLU. 

The training of the color correction model only involves unlabeled natural images captured under good lighting conditions. For each of $n$ natural image in the training set $\{\mathbf{I}_i\}_{i=1}^n$, it is converted into the $L^*a^*b^*$ color space, with the lightness/grayscale values stored in the $L^*$ channel $\{\mathbf{I}_i^{L^*}\}_{i=1}^n$ and the color values stored in the $a^*b^*$ channels $\{\mathbf{I}_i^{a^*b^*}\}_{i=1}^n$. By taking the $L^*$ channel as input and the $a^*b^*$ channels as output, the generator is trained to recover the color channels from the grayscale channel image. Once the training is done, the generator is expected to take the $L^*$ channel of a ``color-corrupted'' image, e.g., taken under artificial lights or poor weather conditions, and produce a natural-looking image as if it was taken under good lighting conditions to achieve the purpose of color correction. The training data for the Discriminator $D$ is the ``real'' original images $\{\mathbf{I}_i\}_{i=1}^n$ and the ``fake'' images formed by concatenating the $L^*$ images and the corresponding $a^*b^*$ images generated by the generator, i.e.,  $\{G(\mathbf{I}^{L^*}_i),\mathbf{I}^{L^*}_i\}_{i=1}^n$. For the training loss $\mathfrak{L}(G,D)$, we use the combination of GAN loss $\mathfrak{L}_{GAN}(G,D)$ and $L1$ loss $\mathfrak{L}_{L1}(G)$ balanced by a weighting factor $\lambda$:
\begin{align}
\mathfrak{L}(G,D)&=\mathfrak{L}_{GAN}(G,D)+\lambda\mathfrak{L}_{L1}(G) \nonumber\\
&=\sum_{i=1}^{n}\log(D(\mathbf{I}_i))+\log(1-D(\{G(\mathbf{I}^{L^*}_i),\mathbf{I}^{L^*}_i\}))\nonumber\\
&+\lambda\sum_{i=1}^{n}\|\mathbf{I}^{a^*b^*}_i-G(\mathbf{I}^{L^*}_i)\|_1
\label{eqn:gan_loss}
\end{align}
The generator $G$ and discriminator $D$ are trained alternatively in an adversarial manner to obtain the final color correction model $G^{\star}{=}\argmin_G\max_D \mathfrak{L}(G,D)$.

\section{Experiments}
\label{sec:experiments}
\begin{table}
\small
\centering
\caption{The details of our Cannabis (Cnbs) dataset and the CVPPP LSC dataset.}
\vspace{2pt}
\begin{tabular}{L{18pt}cC{40pt}C{40pt}C{35pt}C{35pt}l}
\Xhline{0.3ex}
      Dataset & \makecell{Res.\\(pixels)} & \makecell{$\#$ \\training\\images} & \makecell{$\#$ test \\images \\(Natural)} & \makecell{$\#$ test \\images \\(Yellow)} & \makecell{$\#$ test \\images \\(Purple)} & \makecell{Plant \\species}\\  \Xhline{0.2ex}
      Cnbs & $768{\times}768$  & 300  & 40 & 40 & 40 & Cannabis\\ \hline
      A1   & $500{\times}530$  & 128  & 33 & 33 & 33 & Arabidopsis \\ \hline
      A2   & $530{\times}565$  & 31  & 9 & 9 & 9 & Arabidopsis \\ \hline
      A3   & $2448{\times}2048$  & 27  & 65 & 65 & 65 & Tobacco \\ \hline
      A4   & $441{\times}441$  & 624  & 168 & 168 & 168 & Arabidopsis \\ \Xhline{0.3ex}
\end{tabular}
\label{table:datasets}
\end{table}
\begin{figure}[]
\captionsetup[subfigure]{labelformat=empty}
         \centering
         {\fontsize{12pt}{2pt}\selectfont
         \begin{subfigure}{0.04\textwidth}
                 \caption{}
         \end{subfigure}%
         \vspace{-6pt}
         \begin{subfigure}{0.19\textwidth}
                 \caption{Cannabis}
         \end{subfigure}%
         \begin{subfigure}{0.19\textwidth}
                 \caption{A1}
         \end{subfigure}%
         \begin{subfigure}{0.19\textwidth}
                 \caption{A2}
         \end{subfigure}%
         \begin{subfigure}{0.19\textwidth}
                 \caption{A3}
         \end{subfigure}%
         \begin{subfigure}{0.19\textwidth}
                 \caption{A4}
         \end{subfigure}
         }
         \vspace{2pt}
         {\fontsize{12pt}{2pt}\selectfont
         \begin{subfigure}{0.04\textwidth}
         \caption{
            \begin{turn}{90} 
            Natural
            \end{turn}
            }
         \end{subfigure}}%
         \begin{subfigure}{0.19\textwidth}
                 \includegraphics[width=0.95\linewidth]{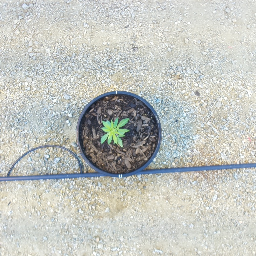}
         \end{subfigure}%
         \begin{subfigure}{0.19\textwidth}
                 \includegraphics[width=0.95\linewidth]{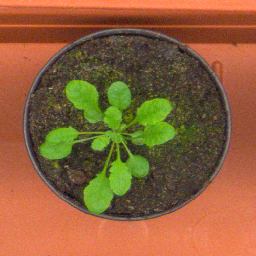}
         \end{subfigure}%
         \begin{subfigure}{0.19\textwidth}
                 \includegraphics[width=0.95\linewidth]{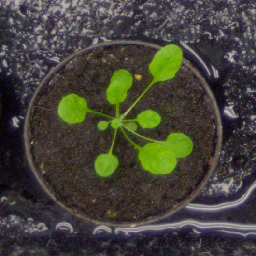}
         \end{subfigure}%
         \begin{subfigure}{0.19\textwidth}
                 \includegraphics[width=0.95\linewidth]{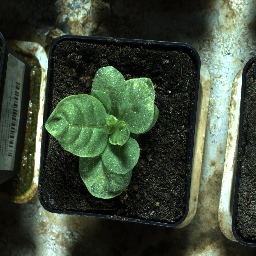}
         \end{subfigure}%
         \begin{subfigure}{0.19\textwidth}
                 \includegraphics[width=0.95\linewidth]{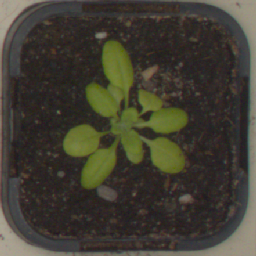}
         \end{subfigure}
         \vspace{2pt} 
         {\fontsize{12pt}{2pt}\selectfont
         \begin{subfigure}{0.04\textwidth}
         \caption{
            \begin{turn}{90} 
            Yellow
            \end{turn}
            }
         \end{subfigure}}%
         \begin{subfigure}{0.19\textwidth}
                 \includegraphics[width=0.95\linewidth]{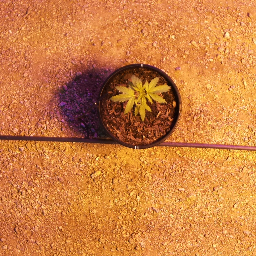}
         \end{subfigure}%
         \begin{subfigure}{0.19\textwidth}
                 \includegraphics[width=0.95\linewidth]{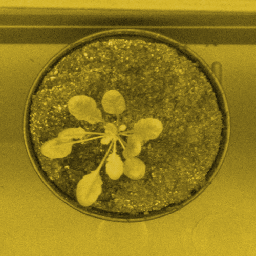}
         \end{subfigure}%
         \begin{subfigure}{0.19\textwidth}
                 \includegraphics[width=0.95\linewidth]{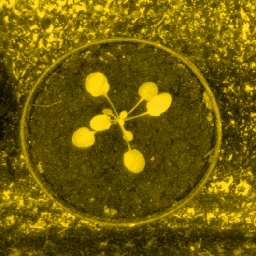}
         \end{subfigure}%
         \begin{subfigure}{0.19\textwidth}
                 \includegraphics[width=0.95\linewidth]{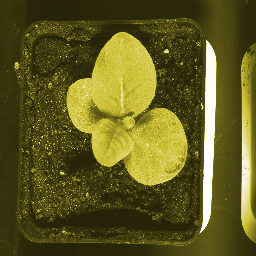}
         \end{subfigure}%
         \begin{subfigure}{0.19\textwidth}
                 \includegraphics[width=0.95\linewidth]{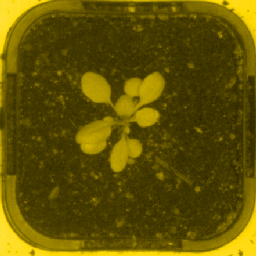}
         \end{subfigure}   
         \vspace{2pt}   
         {\fontsize{12pt}{2pt}\selectfont
         \begin{subfigure}{0.04\textwidth}
         \caption{
            \begin{turn}{90} 
            Purple
            \end{turn}
            }
         \end{subfigure}}%
         \begin{subfigure}{0.19\textwidth}
                 \includegraphics[width=0.95\linewidth]{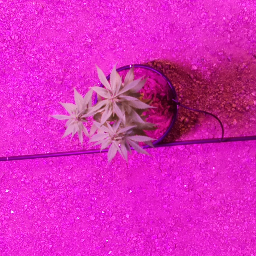}
         \end{subfigure}%
         \begin{subfigure}{0.19\textwidth}
                 \includegraphics[width=0.95\linewidth]{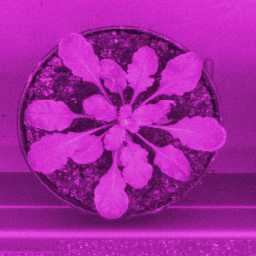}
         \end{subfigure}%
         \begin{subfigure}{0.19\textwidth}
                 \includegraphics[width=0.95\linewidth]{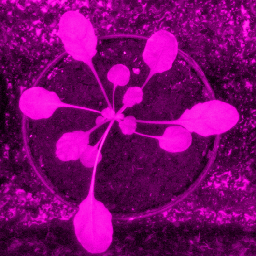}
         \end{subfigure}%
         \begin{subfigure}{0.19\textwidth}
                 \includegraphics[width=0.95\linewidth]{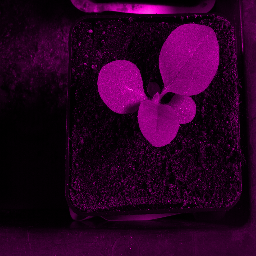}
         \end{subfigure}%
         \begin{subfigure}{0.19\textwidth}
                 \includegraphics[width=0.95\linewidth]{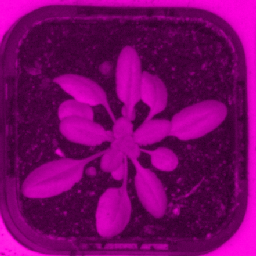}
         \end{subfigure}
         \caption{Examples of ``Natural'', ``Yellow'' and ``Purple'' plant images in our Cannabis dataset and the CVPPP LSC dataset.}
\label{fig:sample_images}
\end{figure}
\subsection{Datasets}
\label{subsec:datasets}
We conduct the experiments on two datasets: our Cannabis dataset and the Computer Vision Problem in Plant Phenotype (CVPPP) Leaf Segmentation Challenge (LSC) dataset \cite{CVPPP_LSC}. Table \ref{table:datasets} summarizes the details of these two datasets. With the image acquisition setup shown in Fig. \ref{fig:image_acquisition_setup}, we collect our Cannabis dataset at different growth stages of cannabis plants under ``Natural'', ``Yellow'', and ``Purple'' lighting conditions, which are controlled by turning off or tuning the grow lights to yellow or purple color. We collect 300 images under the ``Natural'' lighting condition as the unlabeled training set for training our color correction model. Besides, we also collect 40 images under each of the three lighting conditions and manually annotate the leaf segmentation masks for evaluating the leaf segmentation performance. The LSC dataset consists of 4 subsets: A1, A2, A3, and A4, with each subset containing the images of a different plant species. We refer to the original LSC dataset that was acquired under good lighting condition as ``Natural''. To simulate the yellow and purple lighting conditions for the LSC dataset, we generate two more versions, ``Yellow'' and ``Purple'', for each image by randomly adjusting the normalized hue value of each pixel in the ranges of $[0.13, 0.15]$ and $[0.83, 0.86]$, respectively. For both versions, we also randomly adjust the saturation and lightness values in the ranges of $[0.6, 1]$ and $[0.75, 1]$, respectively, to introduce more diversity to the generated images. Sample images under different lighting conditions in our Cannabis dataset and the LSC dataset are shown in Fig. \ref{fig:sample_images}. Note that the training images (without segmentation annotations) in both datasets are only used for the self-supervised training of the color correction model.

\subsection{Training Setup}
\label{subsec:training}
For our self-supervised semantic segmentation model, it does not require any external training data and iteratively updates the segmentation result by resorting to the self-contained information in the same image. For all the experiments, we set the number of semantic labels $K=64$, the maximum iteration number $T=300$ (Algorithm \ref{alg:procedure1}), and the weighting factor $\eta=10$ for the pairwise potential in Eq. (\ref{eqn:gibbs_energy}). While for the color-based leaf segmentation described in Section \ref{subsec:greenness}, we empirically set the mean vector $\vect{\upmu}=[0.3, 0.6, 0.8]$ and covariance matrix $\mathbf{\Sigma}=\text{diag}\left([0.1, 0.3, 0.5]\right)$ of the greenness measurement distribution in Eq. (\ref{eqn:greenness}). The thresholds for the absolute and relative greenness in Eq. (\ref{eqn:threshold}) are set to $\gamma_1=0.2$ and $\gamma_2=0.5$, respectively.

For the self-supervised color correction model, we train two models separately on our Cannabis dataset and the LSC dataset. We divide each image in the training set (without annotation) into blocks of $256{\times}256$ px with $50\%$ overlapping to alleviate the data scarcity issue. We set the weighting factor $\lambda=100$ in Eq. (\ref{eqn:gan_loss}) and train the generator $G$ and the discriminator $D$ alternatively for 50 epoches with batches of size 16. The model weights are updated through Adam optimizer with a learning rate of 0.0002. Due to the large disparity in model complexity between the generator $G$ and the discriminator $D$ in Fig. \ref{fig:flowchart}, training them from the same starting point could easily lead to the earlier convergence of the discriminator. To balance the learning speed of the generator and the discriminator, we pre-train the generator with the $L1$ loss for 20 epoches before alternatively training the two networks. 

\subsection{Evaluation Metrics}
\label{subsec:evaluation}
We evaluate the leaf segmentation performance with the commonly used metric of Foreground-Background Dice (FBD) coefficient, which is calculated as $\text{FBD}{=}\frac{1}{n}\sum_{i=1}^n \frac{2TP_i}{2TP_i+FP_i+FN_i}\in[0,1]$ where $TP_i$, $FP_i$, and $FN_i$ are, respectively, the numbers of true positive, false positive, and false negative pixels of the $i$th image. For the performance evaluation of color correction, we use the metrics of Peak Signal-to-Noise Ratio (PSNR) and Structural Similarity Index (SSIM). A higher PSNR/SSIM value indicates better color correction performance. We tailor the calculation of these two metrics for our color correction task by converting the images to the YUV color space and only measuring PSNR and SSIM of the U and V color channels. Besides, we also employ the Learned Perceptual Image Patch Similarity (LPIPS) metric \cite{zhang2018unreasonable} to measure the perceptual similarity between the original and the color-corrected images. LPIPS between two images is calculated as the distance between their deep embeddings obtained from classic neural networks, e.g., VGG and AlexNet, trained on ImageNet \cite{russakovsky2015imagenet}. 

\subsection{Ablation Studies}
\label{subsec:ablation}
In this subsection, we will examine the influence of different components of our self-supervised leaf segmentation framework using the LSC dataset.

\noindent\textbf{Leaf segmentation.} We first investigate the effects of the smooth term of the fully-connected CRF \cite{krahenbuhl2011efficient}, the absolute greenness and the relative greenness on the performance of leaf segmentation. The results on the ``Natural'' \emph{testing set} of the LSC dataset are shown in Table \ref{table:ablation_segmentation}. The first observation is that including the smooth term negatively contributes to the segmentation results as it gives rise to the chance of removing small leaf regions. The second observation is that using the relative greenness improves the segmentation result on images with green background,  e.g., 94.7\% vs 90.4\% on the subset 'A1' where some images contain green moss in the background. Note that because the relative greenness is a ``relative'' measurement, the non-green regions in an image may have high relative greenness if no green objects present in the image. For this reason, we jointly use the absolute and the relative greenness to reduce false positives caused by only using the relative greenness. 
\begin{table}
\centering
\caption{Analysis of the components of the proposed leaf segmentation algorithm. The highest FBD coefficient is shown in \textbf{bold}. The values in parenthesis are obtained on the subset 'A1'.}
\small
\vspace{2pt}
\begin{tabular}{ccc|c}
\Xhline{0.3ex}
    Smoothness term   & Absolute greenness    & Relative greenness    & FBD(\%)$\uparrow$  \\  \Xhline{0.3ex}
    \OliveGreencheck   & \OliveGreencheck   & \OliveGreencheck          & 94.2 \\ 
    \redcross   & \redcross          & \OliveGreencheck    & 94.6 (A1: 94.7) \\
    \redcross   & \OliveGreencheck   & \redcross           & 95.1 (A1: 90.4) \\
    \redcross   & \OliveGreencheck   & \OliveGreencheck    & \textbf{95.6} (A1: 94.7)  \\ \Xhline{0.3ex}
\end{tabular}
\label{table:ablation_segmentation}
\vspace{-3pt}
\end{table}

\noindent\textbf{Color correction.} For our color correction model, we investigate the effects of three training strategies, including 1) \textbf{Separated}: training on 4 subsets `A1', `A2', `A3', and `A4', separately, 2) \textbf{Combined}: training on the combined dataset of the 4 subsets, and 3) \textbf{Augmented}: training with data augmentations including random flipping, sharpness adjustment, Gaussian blurring, affine transform, and color jittering, to enhance the size and quality of the dataset. Note that for the color jittering augmentation, the hue value of each image remains unchanged to avoid unintended impacts on the color correction task. We train the color correction model on the ``Natural'' (i.e., original) \emph{training set} of the LSC dataset with different training strategies and test the performance on the ``Yellow'' and ``Purple'' \emph{testing sets} of the LSC dataset. The metrics averaged over 4 subsets (A1, A2, A3, and A4) on the `Yellow'' and ``Purple'' \emph{testing sets} are reported in Table \ref{table:ablation_colorization}.
We can see that data augmentations are effective in boosting the color correction performance, while combining the images from different subsets for training provides very limited benefit or may even worsen the performance, e.g., in the case of no augmentations. The worse performance on the combined dataset is probably due to the mutual interference between the notably different image backgrounds and plant traits in different subsets.
\begin{table}
\centering
\caption{Analysis of different training strategies for the color correction model. `$\uparrow$' and `$\downarrow$' respectively indicate that higher or lower values represent better performance. The best results are highlighted with \textcolor{YellowOrange}{\textbf{bold yellow}} (for the ``Yellow'' testing set) and \textcolor{Purple}{\textbf{bold purple}} (for the ``Purple'' testing set) colors.}
\small
\vspace{2pt}
\begin{tabular}{ccc|ccc}
\Xhline{0.3ex}
    Separated     & Combined     & Augmented    & PSNR[dB]$\uparrow$ & SSIM$\uparrow$ & LPIPS$\downarrow$ \\  \Xhline{0.3ex}
    \OliveGreencheck   & \redcross    & \redcross       & \makecell{31.48(Y')\\33.43(P')}             & \makecell{\textcolor{YellowOrange}{\textbf{0.841}}(Y')\\0.860(P')}        & \makecell{0.431(Y')\\0.244(P')}  \\ \hline
    \OliveGreencheck   & \redcross    & \OliveGreencheck       & \makecell{31.33(Y')\\\textcolor{Purple}{\textbf{34.88}}(P')}    & \makecell{0.836(Y')\\\textcolor{Purple}{\textbf{0.872}}(P')}      & \makecell{\textcolor{YellowOrange}{\textbf{0.426}}(Y')\\0.239(P')}  \\ \hline
    \redcross   & \OliveGreencheck    & \redcross       & \makecell{29.54(Y')\\32.46(P')}      & \makecell{0.831(Y')\\0.862(P')}        & \makecell{0.469(Y')\\0.263(P')}\\ \hline
    \redcross   & \OliveGreencheck    & \OliveGreencheck   & \makecell{\textcolor{YellowOrange}{\textbf{32.22}}(Y')\\34.42(P')}   & \makecell{0.831(Y')\\0.860(P')}        & \makecell{0.427(Y')\\\textcolor{Purple}{\textbf{0.237}}(P')}  \\ \Xhline{0.3ex}
\end{tabular}
\label{table:ablation_colorization}
\end{table}

\subsection{Performance Analyses}
\label{subsec:comparison_sota}
\begin{table}
\centering
\caption{Leaf segmentation results in terms of FBD(\%) on the ``Natural'' testing sets of the LSC dataset and our Cannabis dataset for unsupervised (EM \cite{leafsnap}, MCS \cite{MCS}, and Nottingham \cite{collation}), supervised (DC \cite{deepcoloring}, SYN \cite{synthetic}, and UPG \cite{UPGen}), and self-supervised (SSSLIC \cite{kanezaki2018unsupervised}, SSCL \cite{kim2020unsupervised}, and our proposed SSCRF) methods. `$^\star$' indicates that the pre-trained model is fine-tuned on the \emph{training set} of the LSC dataset. The highest FBD coefficient on each dataset/subset is highlighted in \textbf{bold}.}
\small
\vspace{2pt}
\setlength\tabcolsep{3.5pt}\renewcommand\arraystretch{1.25}
  \noindent\makebox[\textwidth]{%
    \begin{tabular}{lcccccccccc}
      \Xhline{0.3ex}
      \multirow{2}{*}{Dataset}
                   & \multicolumn{3}{c}{Unsupervised} & \multicolumn{4}{c}{Supervised} & \multicolumn{3}{c}{Self-supervised} \\ \cmidrule(lr) {2-4} \cmidrule(lr){5-8} \cmidrule(lr){9-11}
                   & EM   & MCS  & Nott.       & DC$^\star$         & SYN  & UPG & UPG$^\star$ & SSSLIC  & SSCL & SSCRF \\  \Xhline{0.2ex}
      Cnbs & 16.1  & 70.6  & 90.1  & --  & 62.2 & 23.0 & --  & 80.7 & 87.8  & \textbf{94.8} \\ \hline
      A1   & 38.5  & 73.6  & 95.3  & 93.3 & 90.3 & 49.2 & 90.4 & 91.5 & 94.3 & \textbf{94.7}  \\ \hline
      A2   & 65.6  & 80.4  & \textbf{93.0}  & 80.3 & 79.3 & 30.8 & 91.0 & 55.8 & 82.4 & 92.0 \\ \hline
      A3   & 34.6  & 39.2  & 90.7  & 68.4 & 72.0 & 36.4 & 92.6 & 91.7 & 93.9 & \textbf{95.2}  \\ \hline
      A4   & 50.2  & 79.2  & 90.2  & 74.7 & 76.8 & 26.4 & 93.2 & 76.2 & 84.7 & \textbf{96.1} \\ \Xhline{0.3ex}
    \end{tabular}
}%
\label{table:natural}
\end{table}

\begin{figure*}[]
\captionsetup[subfigure]{labelformat=empty}
         \centering
         \begin{subfigure}{0.1\textwidth}
                 \vspace{-120pt}
                 \caption{}
                 \vspace{-150pt}
         \end{subfigure}%
         \begin{subfigure}{0.25\textwidth}
                 \vspace{-120pt}
                 \caption{Unsupervised}
                 \vspace{-150pt}
         \end{subfigure}%
         \begin{subfigure}{0.4\textwidth}
                 \vspace{-120pt}
                 \caption{Supervised}
                 \vspace{-150pt}
         \end{subfigure}%
         \begin{subfigure}{0.25\textwidth}
                 \vspace{-120pt}
                 \caption{Self-supervised}
                 \vspace{-150pt}
         \end{subfigure}
         \begin{subfigure}{0.25\textwidth}
                 \vspace{-2pt}
                 $\;\;\overbrace{\qquad\qquad\qquad\qquad}$
                 \caption{}
                 \vspace{-30pt}
         \end{subfigure}%
         \begin{subfigure}{0.3\textwidth}
                 \vspace{-2pt}
                 $\;\;\;\;\;\;\overbrace{\qquad\qquad\qquad\qquad\qquad}$
                 \caption{}
                 \vspace{-30pt}
         \end{subfigure}%
         \begin{subfigure}{0.25\textwidth}
                 \vspace{-2pt}
                 $\quad\quad\;\;\;\;\overbrace{\qquad\qquad\qquad\qquad}$
                 \caption{}
                 \vspace{-30pt}
         \end{subfigure}
                  \begin{subfigure}{0.09\textwidth}
                 \vspace{-2pt}
                 \caption{Image}
                 \vspace{-5pt}
         \end{subfigure}%
         \begin{subfigure}{0.09\textwidth}
                 \vspace{-2pt}
                 \caption{EM}
                 \vspace{-5pt}
         \end{subfigure}%
         \begin{subfigure}{0.09\textwidth}
                 \vspace{-2pt}
                 \caption{\scriptsize{MCS}}
                 \vspace{-6pt}
         \end{subfigure}%
         \begin{subfigure}{0.09\textwidth}
                 \vspace{-2pt}
                 \caption{\scriptsize{Nott.}}
                 \vspace{-6pt}
         \end{subfigure}%
         \begin{subfigure}{0.09\textwidth}
                 \vspace{-2pt}
                 \caption{DC$^\star$}
                 \vspace{-5pt}
         \end{subfigure}%
         \begin{subfigure}{0.09\textwidth}
                 \vspace{-2pt}
                 \caption{\scriptsize{SYN}}
                 \vspace{-6pt}
         \end{subfigure}%
         \begin{subfigure}{0.09\textwidth}
                 \vspace{-2pt}
                 \caption{\scriptsize{UPG}}
                 \vspace{-6pt}
         \end{subfigure}%
         \begin{subfigure}{0.09\textwidth}
                 \vspace{-2pt}
                 \caption{\scriptsize{UPG$^\star$}}
                 \vspace{-6pt}
         \end{subfigure}%
         \begin{subfigure}{0.09\textwidth}
                 \vspace{-2pt}
                 \caption{\scriptsize{SSSLIC}}
                 \vspace{-6pt}
         \end{subfigure}%
         \begin{subfigure}{0.09\textwidth}
                 \vspace{-2pt}
                 \caption{\scriptsize{SSCL}}
                 \vspace{-6pt}
         \end{subfigure}%
         \begin{subfigure}{0.09\textwidth}
                 \vspace{-2pt}
                 \caption{\scriptsize{SSCRF}}
                 \vspace{-6pt}
         \end{subfigure}
         \vspace{1pt}
         \begin{subfigure}{0.09\textwidth}
                 \includegraphics[width=0.98\linewidth]{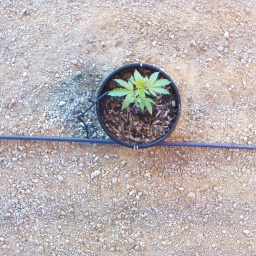}
         \end{subfigure}%
         \begin{subfigure}{0.09\textwidth}
                 \includegraphics[width=0.98\linewidth]{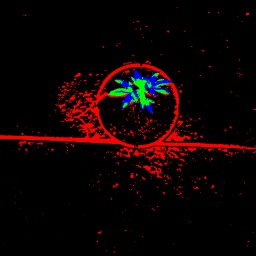}
         \end{subfigure}%
         \begin{subfigure}{0.09\textwidth}
                 \includegraphics[width=0.98\linewidth]{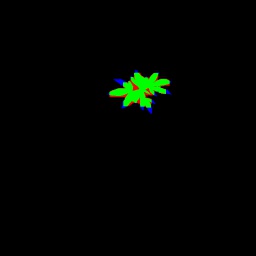}
         \end{subfigure}%
         \begin{subfigure}{0.09\textwidth}
                 \includegraphics[width=0.98\linewidth]{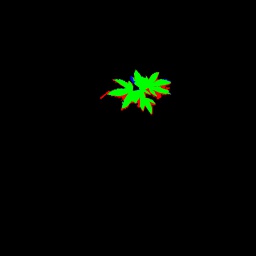}
         \end{subfigure}%
         \begin{subfigure}{0.09\textwidth}
                 \includegraphics[width=0.98\linewidth]{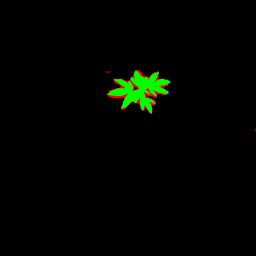}
         \end{subfigure}%
         \begin{subfigure}{0.09\textwidth}
                 \includegraphics[width=0.98\linewidth]{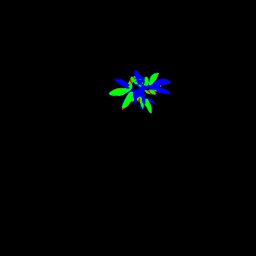}
         \end{subfigure}%
         \begin{subfigure}{0.09\textwidth}
                 \includegraphics[width=0.98\linewidth]{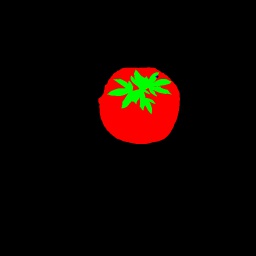}
         \end{subfigure}%
         \begin{subfigure}{0.09\textwidth}
                 \includegraphics[width=0.98\linewidth]{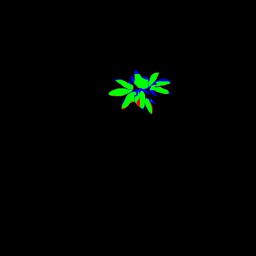}
         \end{subfigure}%
         \begin{subfigure}{0.09\textwidth}
                 \includegraphics[width=0.98\linewidth]{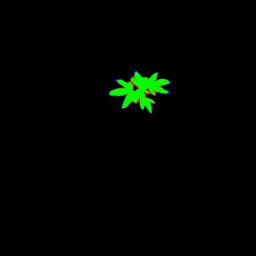}
         \end{subfigure}%
         \begin{subfigure}{0.09\textwidth}
                 \includegraphics[width=0.98\linewidth]{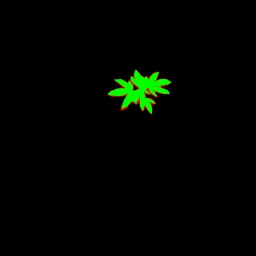}
         \end{subfigure}%
         \begin{subfigure}{0.09\textwidth}
                 \includegraphics[width=0.98\linewidth]{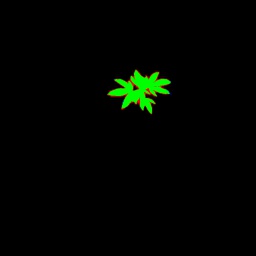}
         \end{subfigure}
         \vspace{1pt}
         \begin{subfigure}{0.09\textwidth}
                 \includegraphics[width=0.98\linewidth]{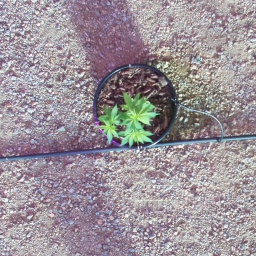}
         \end{subfigure}%
         \begin{subfigure}{0.09\textwidth}
                 \includegraphics[width=0.98\linewidth]{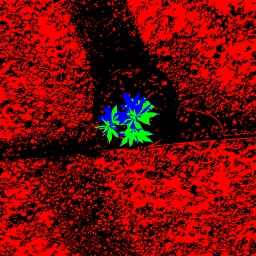}
         \end{subfigure}%
         \begin{subfigure}{0.09\textwidth}
                 \includegraphics[width=0.98\linewidth]{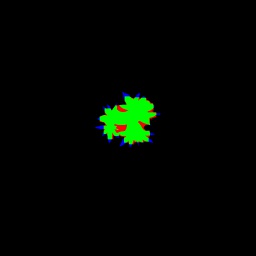}
         \end{subfigure}%
         \begin{subfigure}{0.09\textwidth}
                 \includegraphics[width=0.98\linewidth]{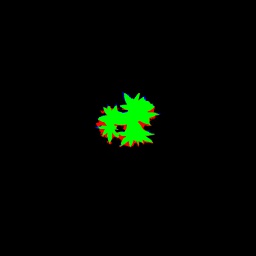}
         \end{subfigure}%
         \begin{subfigure}{0.09\textwidth}
                 \includegraphics[width=0.98\linewidth]{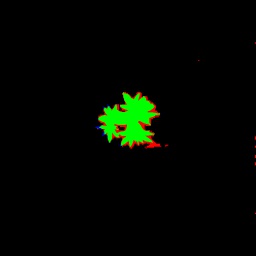}
         \end{subfigure}%
         \begin{subfigure}{0.09\textwidth}
                 \includegraphics[width=0.98\linewidth]{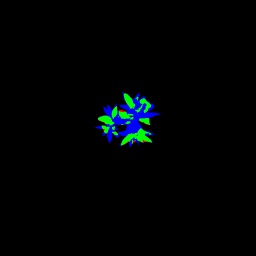}
         \end{subfigure}%
         \begin{subfigure}{0.09\textwidth}
                 \includegraphics[width=0.98\linewidth]{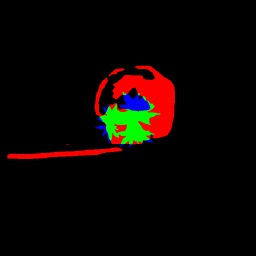}
         \end{subfigure}%
         \begin{subfigure}{0.09\textwidth}
                 \includegraphics[width=0.98\linewidth]{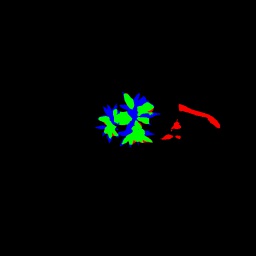}
         \end{subfigure}%
         \begin{subfigure}{0.09\textwidth}
                 \includegraphics[width=0.98\linewidth]{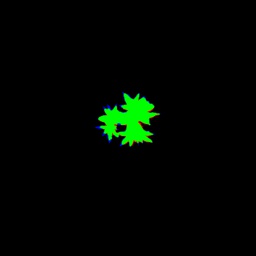}
         \end{subfigure}%
         \begin{subfigure}{0.09\textwidth}
                 \includegraphics[width=0.98\linewidth]{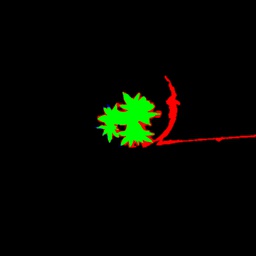}
         \end{subfigure}%
         \begin{subfigure}{0.09\textwidth}
                 \includegraphics[width=0.98\linewidth]{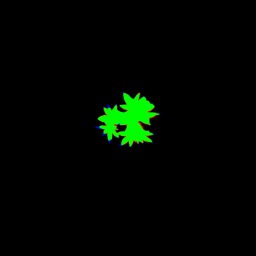}
         \end{subfigure}
         \vspace{1pt}
         \begin{subfigure}{0.09\textwidth}
                 \includegraphics[width=0.98\linewidth]{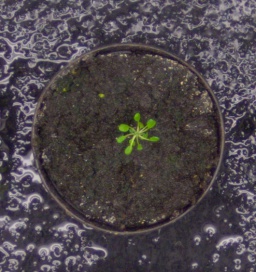}
         \end{subfigure}%
         \begin{subfigure}{0.09\textwidth}
                 \includegraphics[width=0.98\linewidth]{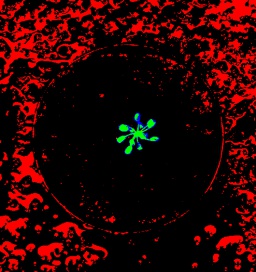}
         \end{subfigure}%
         \begin{subfigure}{0.09\textwidth}
                 \includegraphics[width=0.98\linewidth]{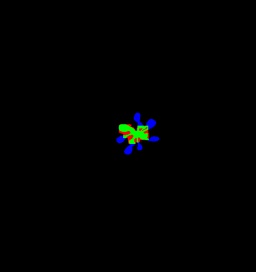}
         \end{subfigure}%
         \begin{subfigure}{0.09\textwidth}
                 \includegraphics[width=0.98\linewidth]{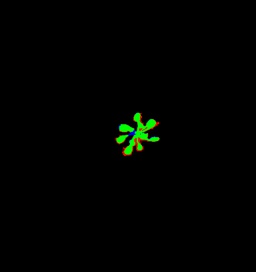}
         \end{subfigure}%
         \begin{subfigure}{0.09\textwidth}
                 \includegraphics[width=0.98\linewidth]{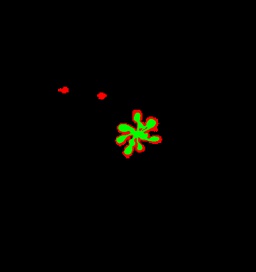}
         \end{subfigure}%
         \begin{subfigure}{0.09\textwidth}
                 \includegraphics[width=0.98\linewidth]{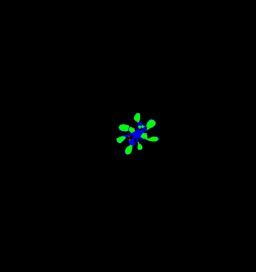}
         \end{subfigure}%
         \begin{subfigure}{0.09\textwidth}
                 \includegraphics[width=0.98\linewidth]{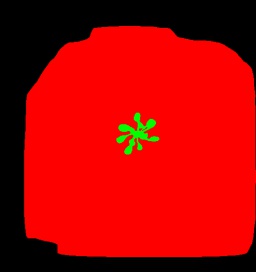}
         \end{subfigure}%
         \begin{subfigure}{0.09\textwidth}
                 \includegraphics[width=0.98\linewidth]{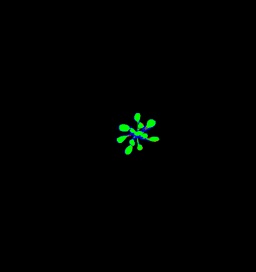}
         \end{subfigure}%
         \begin{subfigure}{0.09\textwidth}
                 \includegraphics[width=0.98\linewidth]{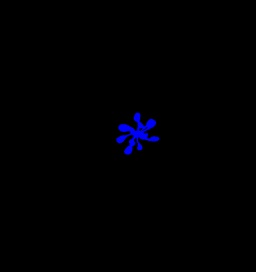}
         \end{subfigure}%
         \begin{subfigure}{0.09\textwidth}
                 \includegraphics[width=0.98\linewidth]{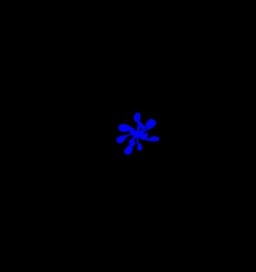}
         \end{subfigure}%
         \begin{subfigure}{0.09\textwidth}
                 \includegraphics[width=0.98\linewidth]{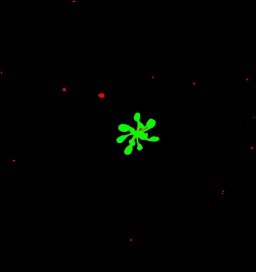}
         \end{subfigure}
         \vspace{1pt}
         \begin{subfigure}{0.09\textwidth}
                 \includegraphics[width=0.98\linewidth]{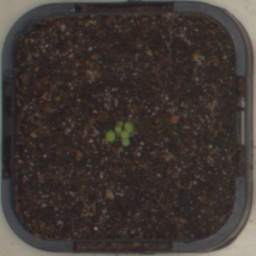}
         \end{subfigure}%
         \begin{subfigure}{0.09\textwidth}
                 \includegraphics[width=0.98\linewidth]{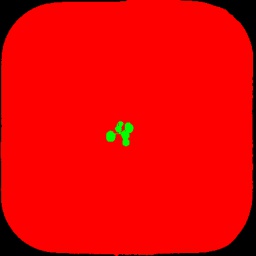}
         \end{subfigure}%
         \begin{subfigure}{0.09\textwidth}
                 \includegraphics[width=0.98\linewidth]{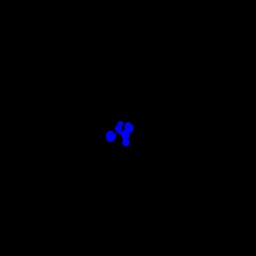}
         \end{subfigure}%
         \begin{subfigure}{0.09\textwidth}
                 \includegraphics[width=0.98\linewidth]{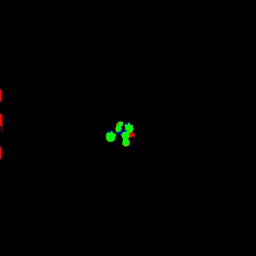}
         \end{subfigure}%
         \begin{subfigure}{0.09\textwidth}
                 \includegraphics[width=0.98\linewidth]{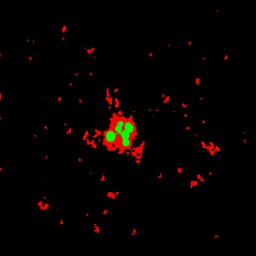}
         \end{subfigure}%
         \begin{subfigure}{0.09\textwidth}
                 \includegraphics[width=0.98\linewidth]{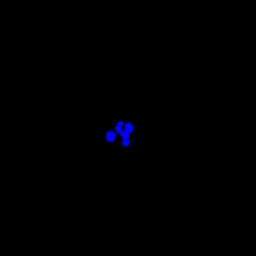}
         \end{subfigure}%
         \begin{subfigure}{0.09\textwidth}
                 \includegraphics[width=0.98\linewidth]{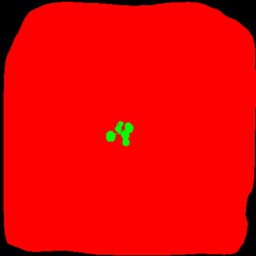}
         \end{subfigure}%
         \begin{subfigure}{0.09\textwidth}
                 \includegraphics[width=0.98\linewidth]{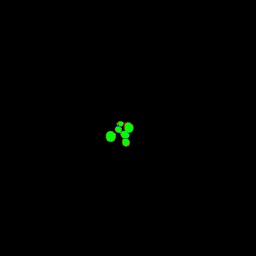}
         \end{subfigure}%
         \begin{subfigure}{0.09\textwidth}
                 \includegraphics[width=0.98\linewidth]{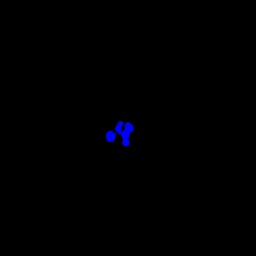}
         \end{subfigure}%
         \begin{subfigure}{0.09\textwidth}
                 \includegraphics[width=0.98\linewidth]{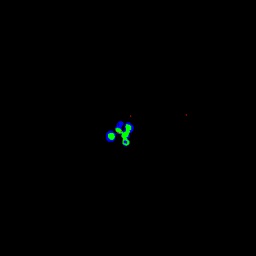}
         \end{subfigure}%
         \begin{subfigure}{0.09\textwidth}
                 \includegraphics[width=0.98\linewidth]{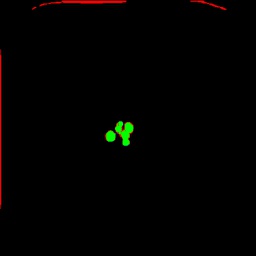}
         \end{subfigure}
         \vspace{1pt}
         \begin{subfigure}{0.09\textwidth}
                 \includegraphics[width=0.98\linewidth]{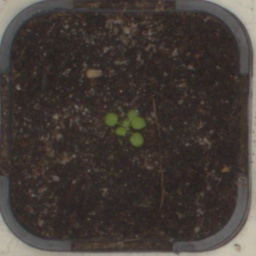}
         \end{subfigure}%
         \begin{subfigure}{0.09\textwidth}
                 \includegraphics[width=0.98\linewidth]{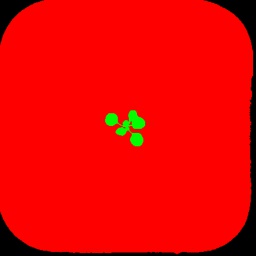}
         \end{subfigure}%
         \begin{subfigure}{0.09\textwidth}
                 \includegraphics[width=0.98\linewidth]{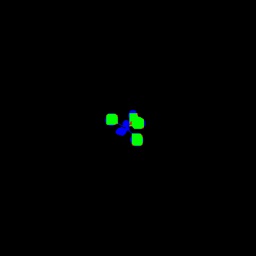}
         \end{subfigure}%
         \begin{subfigure}{0.09\textwidth}
                 \includegraphics[width=0.98\linewidth]{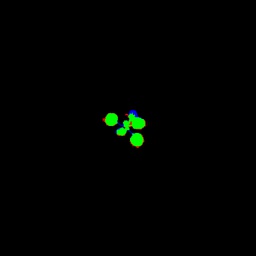}
         \end{subfigure}%
         \begin{subfigure}{0.09\textwidth}
                 \includegraphics[width=0.98\linewidth]{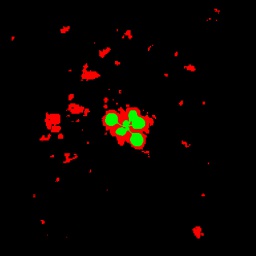}
         \end{subfigure}%
         \begin{subfigure}{0.09\textwidth}
                 \includegraphics[width=0.98\linewidth]{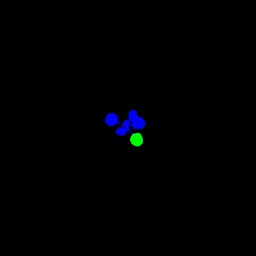}
         \end{subfigure}%
         \begin{subfigure}{0.09\textwidth}
                 \includegraphics[width=0.98\linewidth]{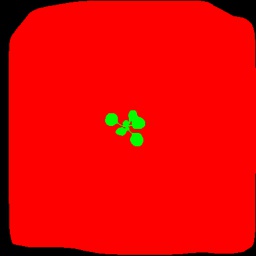}
         \end{subfigure}%
         \begin{subfigure}{0.09\textwidth}
                 \includegraphics[width=0.98\linewidth]{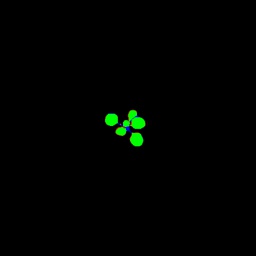}
         \end{subfigure}%
         \begin{subfigure}{0.09\textwidth}
                 \includegraphics[width=0.98\linewidth]{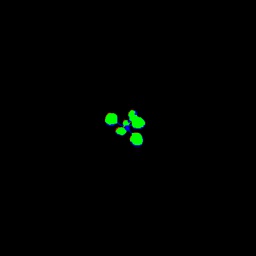}
         \end{subfigure}%
         \begin{subfigure}{0.09\textwidth}
                 \includegraphics[width=0.98\linewidth]{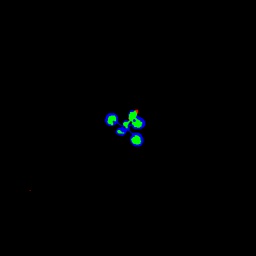}
         \end{subfigure}%
         \begin{subfigure}{0.09\textwidth}
                 \includegraphics[width=0.98\linewidth]{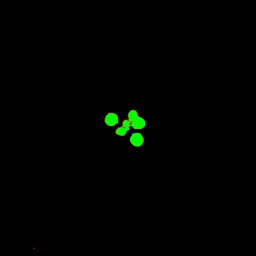}
         \end{subfigure}
                  \vspace{1pt}
         \begin{subfigure}{0.09\textwidth}
                 \includegraphics[width=0.98\linewidth]{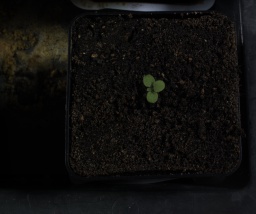}
         \end{subfigure}%
         \begin{subfigure}{0.09\textwidth}
                 \includegraphics[width=0.98\linewidth]{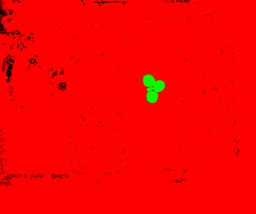}
         \end{subfigure}%
         \begin{subfigure}{0.09\textwidth}
                 \includegraphics[width=0.98\linewidth]{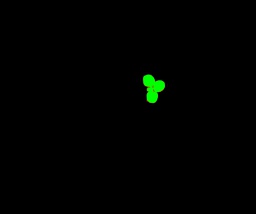}
         \end{subfigure}%
         \begin{subfigure}{0.09\textwidth}
                 \includegraphics[width=0.98\linewidth]{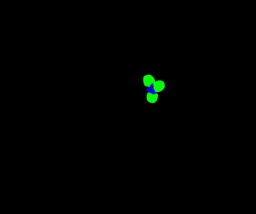}
         \end{subfigure}%
         \begin{subfigure}{0.09\textwidth}
                 \includegraphics[width=0.98\linewidth]{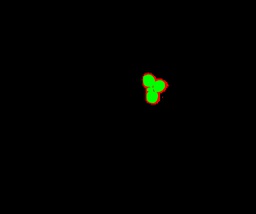}
         \end{subfigure}%
         \begin{subfigure}{0.09\textwidth}
                 \includegraphics[width=0.98\linewidth]{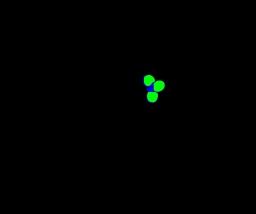}
         \end{subfigure}%
         \begin{subfigure}{0.09\textwidth}
                 \includegraphics[width=0.98\linewidth]{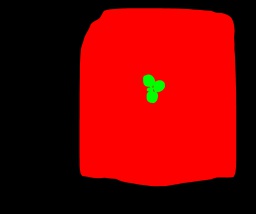}
         \end{subfigure}%
         \begin{subfigure}{0.09\textwidth}
                 \includegraphics[width=0.98\linewidth]{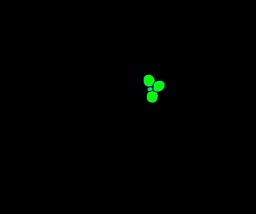}
         \end{subfigure}%
         \begin{subfigure}{0.09\textwidth}
                 \includegraphics[width=0.98\linewidth]{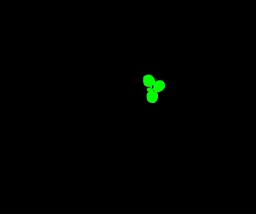}
         \end{subfigure}%
         \begin{subfigure}{0.09\textwidth}
                 \includegraphics[width=0.98\linewidth]{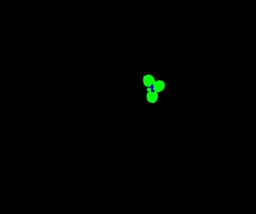}
         \end{subfigure}%
         \begin{subfigure}{0.09\textwidth}
                 \includegraphics[width=0.98\linewidth]{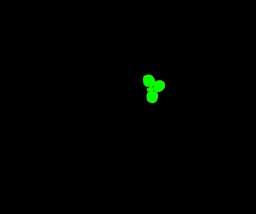}
         \end{subfigure}
         \vspace{1pt}
         \begin{subfigure}{0.09\textwidth}
                 \includegraphics[width=0.98\linewidth]{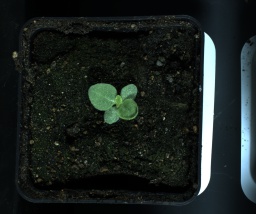}
         \end{subfigure}%
         \begin{subfigure}{0.09\textwidth}
                 \includegraphics[width=0.98\linewidth]{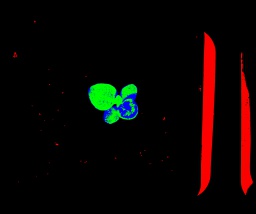}
         \end{subfigure}%
         \begin{subfigure}{0.09\textwidth}
                 \includegraphics[width=0.98\linewidth]{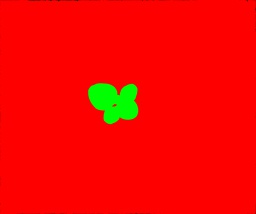}
         \end{subfigure}%
         \begin{subfigure}{0.09\textwidth}
                 \includegraphics[width=0.98\linewidth]{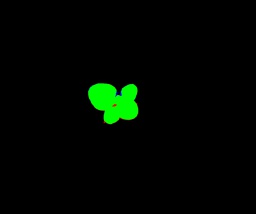}
         \end{subfigure}%
         \begin{subfigure}{0.09\textwidth}
                 \includegraphics[width=0.98\linewidth]{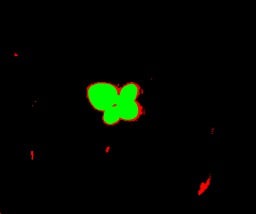}
         \end{subfigure}%
         \begin{subfigure}{0.09\textwidth}
                 \includegraphics[width=0.98\linewidth]{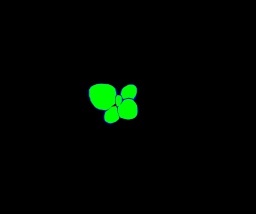}
         \end{subfigure}%
         \begin{subfigure}{0.09\textwidth}
                 \includegraphics[width=0.98\linewidth]{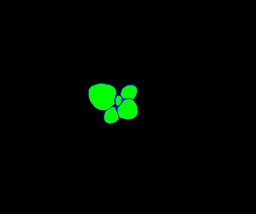}
         \end{subfigure}%
         \begin{subfigure}{0.09\textwidth}
                 \includegraphics[width=0.98\linewidth]{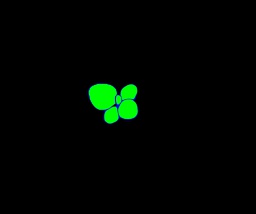}
         \end{subfigure}%
         \begin{subfigure}{0.09\textwidth}
                 \includegraphics[width=0.98\linewidth]{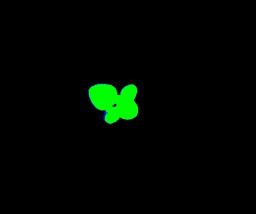}
         \end{subfigure}%
         \begin{subfigure}{0.09\textwidth}
                 \includegraphics[width=0.98\linewidth]{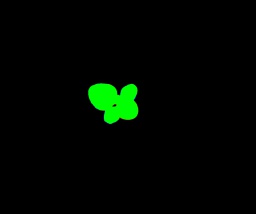}
         \end{subfigure}%
         \begin{subfigure}{0.09\textwidth}
                 \includegraphics[width=0.98\linewidth]{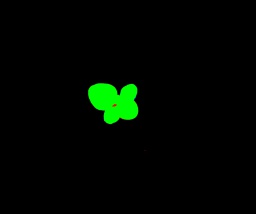}
         \end{subfigure}
         \vspace{1pt}
         \begin{subfigure}{0.09\textwidth}
                 \includegraphics[width=0.98\linewidth]{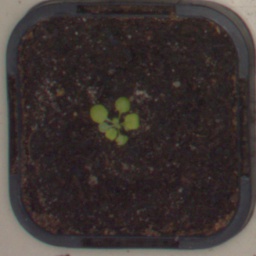}
         \end{subfigure}%
         \begin{subfigure}{0.09\textwidth}
                 \includegraphics[width=0.98\linewidth]{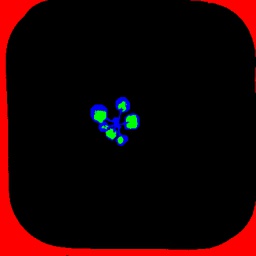}
         \end{subfigure}%
         \begin{subfigure}{0.09\textwidth}
                 \includegraphics[width=0.98\linewidth]{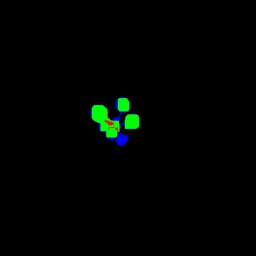}
         \end{subfigure}%
         \begin{subfigure}{0.09\textwidth}
                 \includegraphics[width=0.98\linewidth]{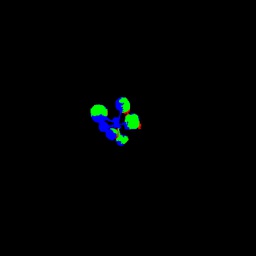}
         \end{subfigure}%
         \begin{subfigure}{0.09\textwidth}
                 \includegraphics[width=0.98\linewidth]{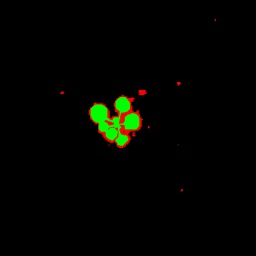}
         \end{subfigure}%
         \begin{subfigure}{0.09\textwidth}
                 \includegraphics[width=0.98\linewidth]{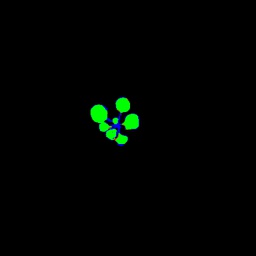}
         \end{subfigure}%
         \begin{subfigure}{0.09\textwidth}
                 \includegraphics[width=0.98\linewidth]{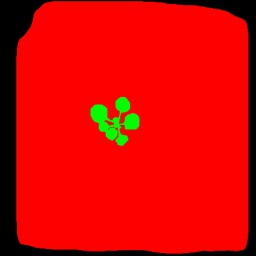}
         \end{subfigure}%
         \begin{subfigure}{0.09\textwidth}
                 \includegraphics[width=0.98\linewidth]{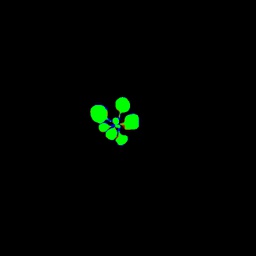}
         \end{subfigure}%
         \begin{subfigure}{0.09\textwidth}
                 \includegraphics[width=0.98\linewidth]{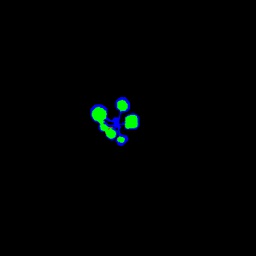}
         \end{subfigure}%
         \begin{subfigure}{0.09\textwidth}
                 \includegraphics[width=0.98\linewidth]{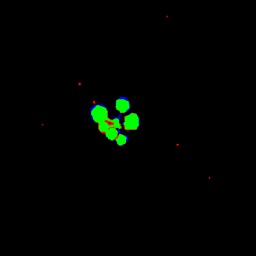}
         \end{subfigure}%
         \begin{subfigure}{0.09\textwidth}
                 \includegraphics[width=0.98\linewidth]{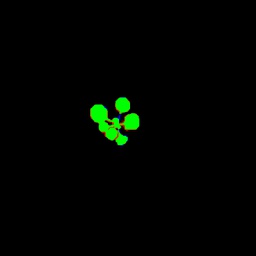}
         \end{subfigure}
         \vspace{1pt}
         \begin{subfigure}{0.09\textwidth}
                 \includegraphics[width=0.98\linewidth]{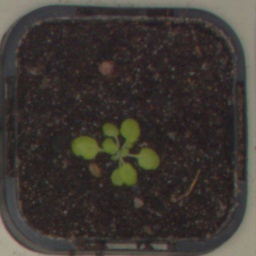}
         \end{subfigure}%
         \begin{subfigure}{0.09\textwidth}
                 \includegraphics[width=0.98\linewidth]{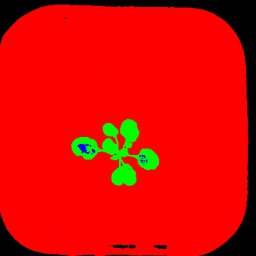}
         \end{subfigure}%
         \begin{subfigure}{0.09\textwidth}
                 \includegraphics[width=0.98\linewidth]{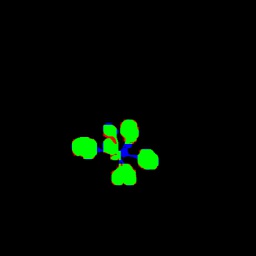}
         \end{subfigure}%
         \begin{subfigure}{0.09\textwidth}
                 \includegraphics[width=0.98\linewidth]{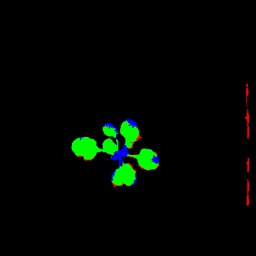}
         \end{subfigure}%
         \begin{subfigure}{0.09\textwidth}
                 \includegraphics[width=0.98\linewidth]{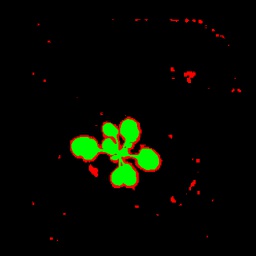}
         \end{subfigure}%
         \begin{subfigure}{0.09\textwidth}
                 \includegraphics[width=0.98\linewidth]{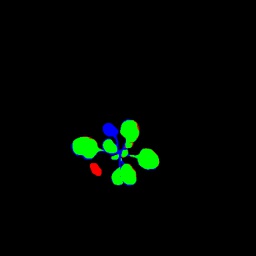}
         \end{subfigure}%
         \begin{subfigure}{0.09\textwidth}
                 \includegraphics[width=0.98\linewidth]{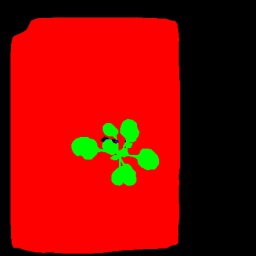}
         \end{subfigure}%
         \begin{subfigure}{0.09\textwidth}
                 \includegraphics[width=0.98\linewidth]{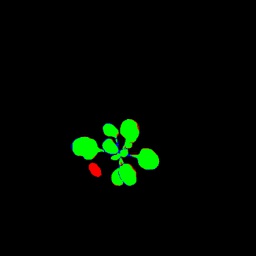}
         \end{subfigure}%
         \begin{subfigure}{0.09\textwidth}
                 \includegraphics[width=0.98\linewidth]{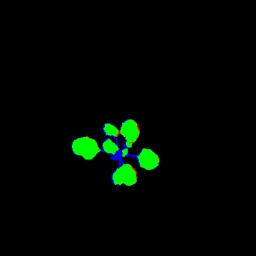}
         \end{subfigure}%
         \begin{subfigure}{0.09\textwidth}
                 \includegraphics[width=0.98\linewidth]{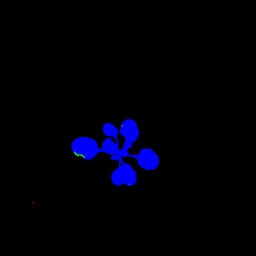}
         \end{subfigure}%
         \begin{subfigure}{0.09\textwidth}
                 \includegraphics[width=0.98\linewidth]{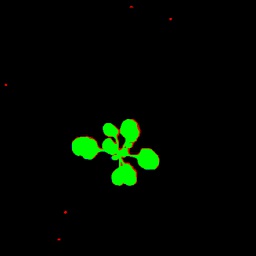}
         \end{subfigure}
         \begin{subfigure}{0.09\textwidth}
                 \includegraphics[width=0.98\linewidth]{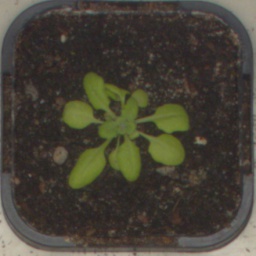}
         \end{subfigure}%
         \begin{subfigure}{0.09\textwidth}
                 \includegraphics[width=0.98\linewidth]{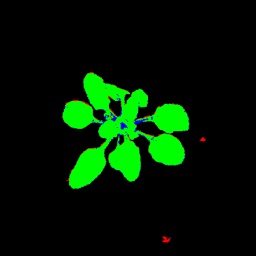}
         \end{subfigure}%
         \begin{subfigure}{0.09\textwidth}
                 \includegraphics[width=0.98\linewidth]{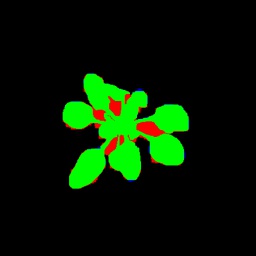}
         \end{subfigure}%
         \begin{subfigure}{0.09\textwidth}
                 \includegraphics[width=0.98\linewidth]{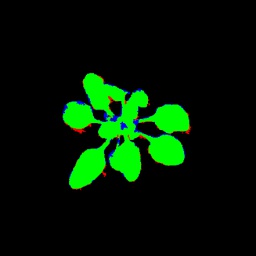}
         \end{subfigure}%
         \begin{subfigure}{0.09\textwidth}
                 \includegraphics[width=0.98\linewidth]{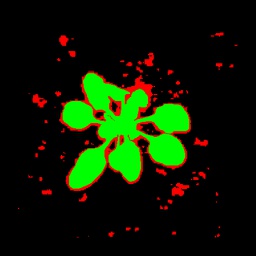}
         \end{subfigure}%
         \begin{subfigure}{0.09\textwidth}
                 \includegraphics[width=0.98\linewidth]{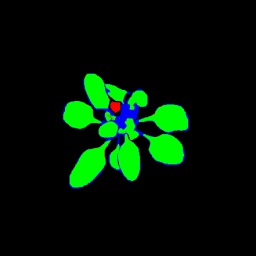}
         \end{subfigure}%
         \begin{subfigure}{0.09\textwidth}
                 \includegraphics[width=0.98\linewidth]{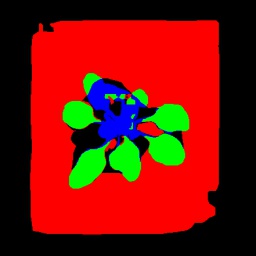}
         \end{subfigure}%
         \begin{subfigure}{0.09\textwidth}
                 \includegraphics[width=0.98\linewidth]{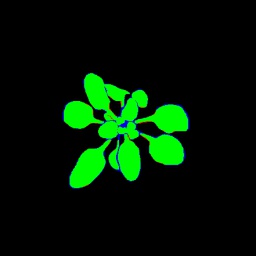}
         \end{subfigure}%
         \begin{subfigure}{0.09\textwidth}
                 \includegraphics[width=0.98\linewidth]{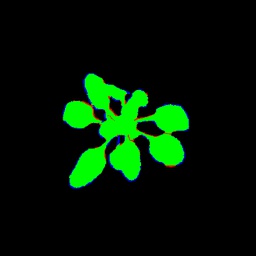}
         \end{subfigure}%
         \begin{subfigure}{0.09\textwidth}
                 \includegraphics[width=0.98\linewidth]{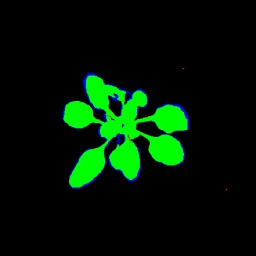}
         \end{subfigure}%
         \begin{subfigure}{0.09\textwidth}
                 \includegraphics[width=0.98\linewidth]{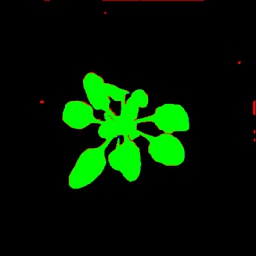}
         \end{subfigure}
         \vspace{1pt}
         \begin{subfigure}{0.09\textwidth}
                 \includegraphics[width=0.98\linewidth]{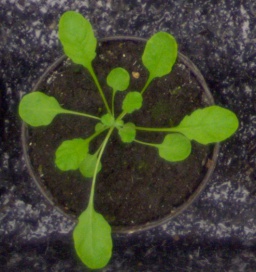}
         \end{subfigure}%
         \begin{subfigure}{0.09\textwidth}
                 \includegraphics[width=0.98\linewidth]{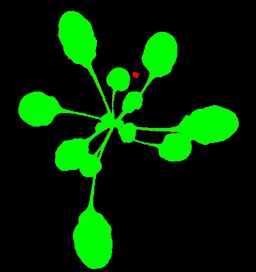}
         \end{subfigure}%
         \begin{subfigure}{0.09\textwidth}
                 \includegraphics[width=0.98\linewidth]{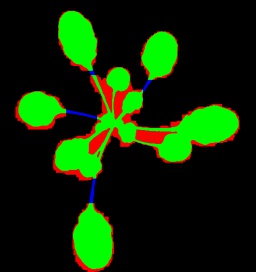}
         \end{subfigure}%
         \begin{subfigure}{0.09\textwidth}
                 \includegraphics[width=0.98\linewidth]{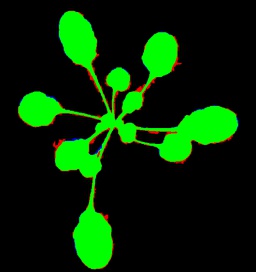}
         \end{subfigure}%
         \begin{subfigure}{0.09\textwidth}
                 \includegraphics[width=0.98\linewidth]{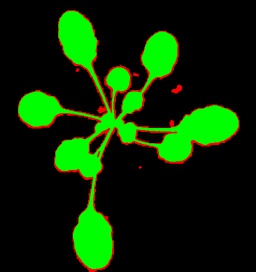}
         \end{subfigure}%
         \begin{subfigure}{0.09\textwidth}
                 \includegraphics[width=0.98\linewidth]{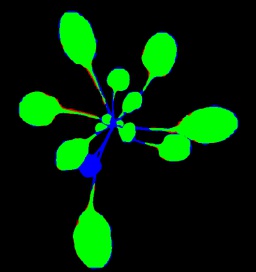}
         \end{subfigure}%
         \begin{subfigure}{0.09\textwidth}
                 \includegraphics[width=0.98\linewidth]{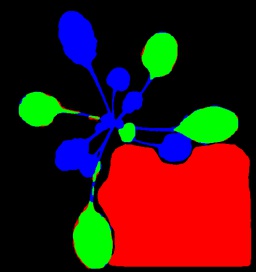}
         \end{subfigure}%
         \begin{subfigure}{0.09\textwidth}
                 \includegraphics[width=0.98\linewidth]{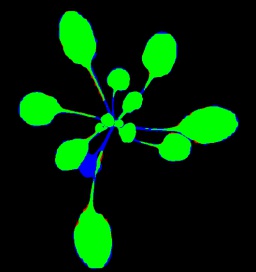}
         \end{subfigure}%
         \begin{subfigure}{0.09\textwidth}
                 \includegraphics[width=0.98\linewidth]{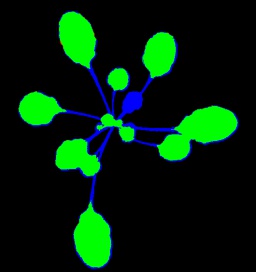}
         \end{subfigure}%
         \begin{subfigure}{0.09\textwidth}
                 \includegraphics[width=0.98\linewidth]{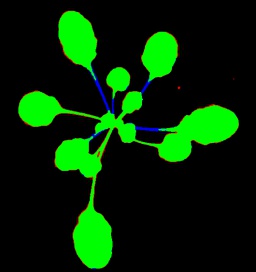}
         \end{subfigure}%
         \begin{subfigure}{0.09\textwidth}
                 \includegraphics[width=0.98\linewidth]{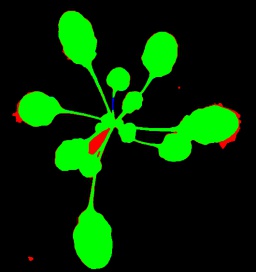}
         \end{subfigure}
         \vspace{1pt}
         \begin{subfigure}{0.09\textwidth}
                 \includegraphics[width=0.98\linewidth]{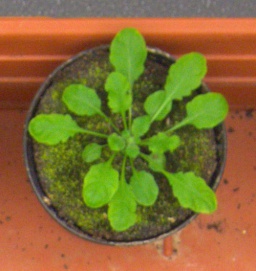}
         \end{subfigure}%
         \begin{subfigure}{0.09\textwidth}
                 \includegraphics[width=0.98\linewidth]{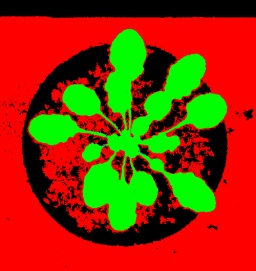}
         \end{subfigure}%
         \begin{subfigure}{0.09\textwidth}
                 \includegraphics[width=0.98\linewidth]{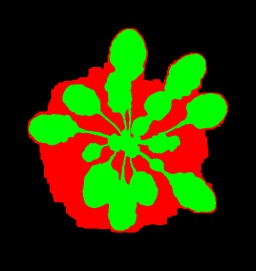}
         \end{subfigure}%
         \begin{subfigure}{0.09\textwidth}
                 \includegraphics[width=0.98\linewidth]{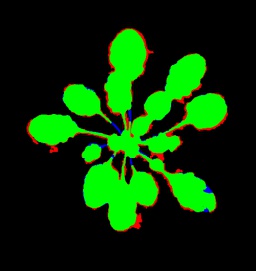}
         \end{subfigure}%
         \begin{subfigure}{0.09\textwidth}
                 \includegraphics[width=0.98\linewidth]{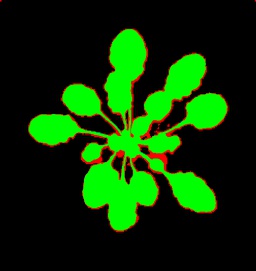}
         \end{subfigure}%
         \begin{subfigure}{0.09\textwidth}
                 \includegraphics[width=0.98\linewidth]{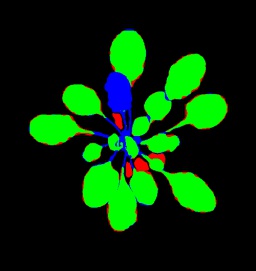}
         \end{subfigure}%
         \begin{subfigure}{0.09\textwidth}
                 \includegraphics[width=0.98\linewidth]{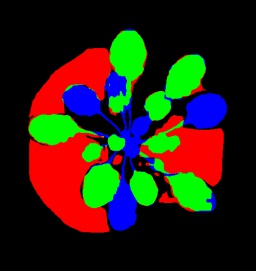}
         \end{subfigure}%
         \begin{subfigure}{0.09\textwidth}
                 \includegraphics[width=0.98\linewidth]{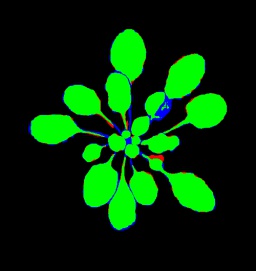}
         \end{subfigure}%
         \begin{subfigure}{0.09\textwidth}
                 \includegraphics[width=0.98\linewidth]{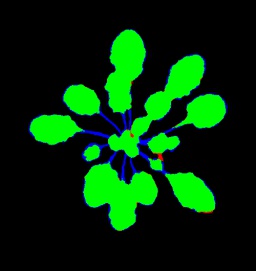}
         \end{subfigure}%
         \begin{subfigure}{0.09\textwidth}
                 \includegraphics[width=0.98\linewidth]{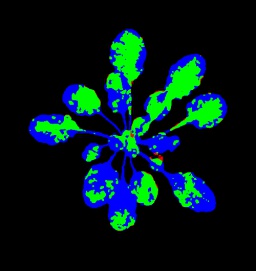}
         \end{subfigure}%
         \begin{subfigure}{0.09\textwidth}
                 \includegraphics[width=0.98\linewidth]{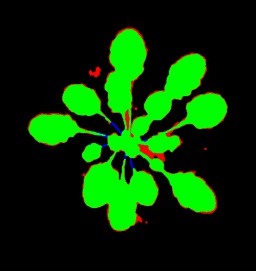}
         \end{subfigure}
         \vspace{1pt}
         \begin{subfigure}{0.09\textwidth}
                 \includegraphics[width=0.98\linewidth]{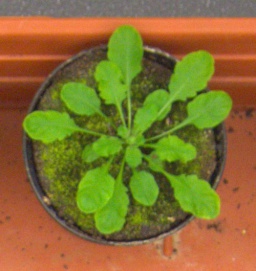}
         \end{subfigure}%
         \begin{subfigure}{0.09\textwidth}
                 \includegraphics[width=0.98\linewidth]{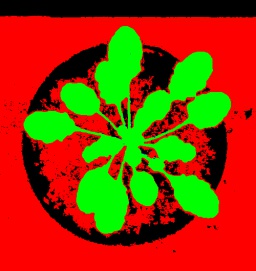}
         \end{subfigure}%
         \begin{subfigure}{0.09\textwidth}
                 \includegraphics[width=0.98\linewidth]{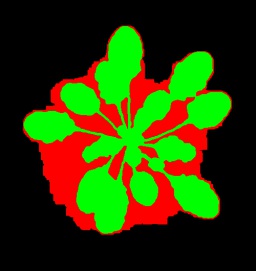}
         \end{subfigure}%
         \begin{subfigure}{0.09\textwidth}
                 \includegraphics[width=0.98\linewidth]{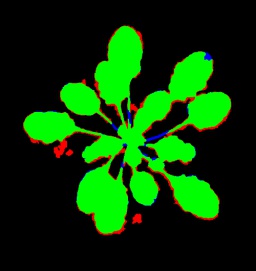}
         \end{subfigure}%
         \begin{subfigure}{0.09\textwidth}
                 \includegraphics[width=0.98\linewidth]{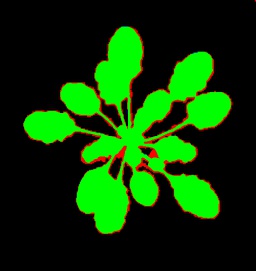}
         \end{subfigure}%
         \begin{subfigure}{0.09\textwidth}
                 \includegraphics[width=0.98\linewidth]{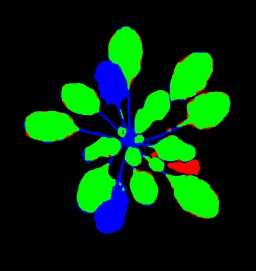}
         \end{subfigure}%
         \begin{subfigure}{0.09\textwidth}
                 \includegraphics[width=0.98\linewidth]{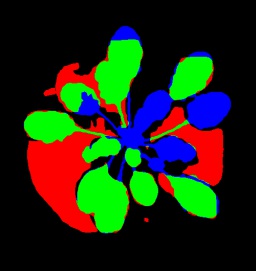}
         \end{subfigure}%
         \begin{subfigure}{0.09\textwidth}
                 \includegraphics[width=0.98\linewidth]{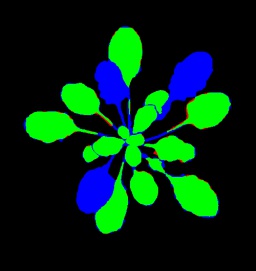}
         \end{subfigure}%
         \begin{subfigure}{0.09\textwidth}
                 \includegraphics[width=0.98\linewidth]{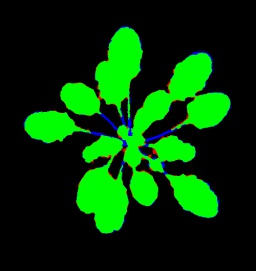}
         \end{subfigure}%
         \begin{subfigure}{0.09\textwidth}
                 \includegraphics[width=0.98\linewidth]{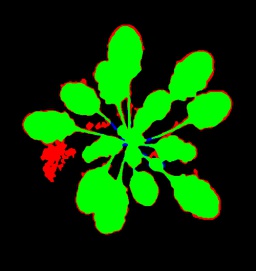}
         \end{subfigure}%
         \begin{subfigure}{0.09\textwidth}
                 \includegraphics[width=0.98\linewidth]{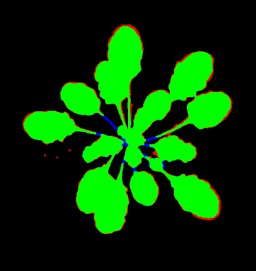}
         \end{subfigure}
         \caption{Example leaf segmentation results. For the results of DC$^\star$\cite{deepcoloring} and UPG$^\star$\cite{UPGen} on the Cannabis dataset, we show the visualization results obtained with the models fine-tuned on the subsets `A2' and `A4', respectively, which offer the highest achievable FBD(\%) metrics according to Table \ref{table:cross_dataset}. Color coding: \textcolor{OliveGreen}{\emph{green}}: detected leaf regions (true positives); \textcolor{red}{\emph{red}}: detected non-leaf regions (false positives); \textcolor{blue}{\emph{blue}}: mis-detected leaf regions (false negatives).}
\label{fig:leaf_segmentation_examples}
\vspace{-1pt}
\end{figure*}

We compare our proposed self-supervised leaf segmentation framework with a wide range of methods covering the categories of unsupervised, supervised and self-supervised methods. The segmentation results on the ``Natural'' \emph{testing sets} of the LSC dataset and our Cannabis dataset are shown in Table \ref{table:natural} and the visualization results on some example leaf images are provided in Fig. \ref{fig:leaf_segmentation_examples}. 

For unsupervised leaf segmentation methods, EM \cite{leafsnap} performs poorly on both datasets as it assumes that the foreground and the background pixels can be modeled with two well-separated Gaussian distributions in the HSV color space, which does not hold for the images in the LSC dataset and our Cannabis dataset. MCS \cite{MCS} incorporates multiple cues, including color, texture, shape, and structure (i.e., leaf vein) information, to facilitate the leaf segmentation in complex backgrounds. However, it lacks the capability of discriminating the leaf from the scattered moss regions and is sensitive to illumination changes, leading to the poor segmentation performance on the subsets `A1' and `A3' of the LSC dataset. Nottingham \cite{collation} first over-segments the image in the $L^*a^*b^*$ color space using the SLIC \cite{achanta2012slic} superpixel algorithm and extracts leaf regions with a simple seeded region growing algorithm in the superpixel space. Despite its great performance, we found that due to the pixel intensity variations among the superpixels within the leaf regions, it is tricky to select an appropriate threshold for separating the leaf regions from the background. Different from Nottingham \cite{collation}, our proposed method performs leaf segmentation at a higher granularity level of semantic objects, thus reducing the intensity inhomogeneity within leaf and non-leaf regions and providing greater flexibility in threshold selection. 

\begin{table}
\centering
\caption{Cross-dataset performance evaluation in terms of FBD(\%) on the ``Natural'' set of our Cannabis dataset for DC$^\star$\cite{deepcoloring} and UPG$^\star$\cite{UPGen}.}
\small
\vspace{2pt}
\setlength\tabcolsep{3.5pt}\renewcommand\arraystretch{1.25}
  \noindent\makebox[\textwidth]{%
    \begin{tabular}{lccccccccc}
      \Xhline{0.3ex}
      \multirow{2}{*}{\makecell{Fine-tuning \\dataset}}     & \multicolumn{4}{c}{DC$^\star$\cite{deepcoloring}} & & \multicolumn{4}{c}{UPG$^\star$\cite{UPGen}} \\ \cline{2-5} \cline{7-10}
                                                            & A1    & A2     & A3     & A4      &     & A1    & A2     & A3     & A4   \\  \Xhline{0.2ex}
      FBD(\%)                                               & 42.6  & 81.3   & 76.5   & 66.5    &     & 74.2  & 81.6   & 53.2   & 83.2 \\ \Xhline{0.3ex}
    \end{tabular}
}%
\label{table:cross_dataset}
\end{table}
We investigate several supervised leaf segmentation methods based on two mainstream segmentation networks, U-Net \cite{Unet} and Mask-RCNN \cite{he2017mask}. By formulating instance-level segmentation as a coloring problem with a fixed number of colors, Deep Coloring (DC) \cite{deepcoloring} allows to train a semantic segmentation network based on U-Net \cite{Unet} for instance-level segmentation with standard semantic segmentation objectives. With the pre-trained model provided by the authors of \cite{deepcoloring}, we fine-tune it on each subset of the \emph{training set} of the LSC dataset and test the fine-tuned model on the corresponding subset of the \emph{testing set} of the LSC dataset. As the first color is reserved to represent the background, we conveniently extract the first network output channel as the leaf segmentation mask. From the visualization results in Fig. \ref{fig:sample_images}, we can see that DC \cite{deepcoloring} gives reasonably good leaf segmentation performance, but it tends to mis-identify salient non-leaf regions in the background as leaf regions. SYN \cite{synthetic} and UPG \cite{UPGen} are methods based on Mask-RCNN \cite{he2017mask} and make use of large-scale synthetic training data to achieve state-of-the-art leaf segmentation performance. For both methods, we use the union of the instance segmentation mask as the final leaf segmentation result. For SYN \cite{synthetic}, we only report the results obtained with the pre-trained model as the implementation of model training has been removed in the source code published by the authors. While for UPG \cite{UPGen}, we report the results obtained with the pre-trained model and the model fine-tuned on each subset of the \emph{training set} of the LSC dataset. We use `$^\star$' to indicate that a model has been fine-tuned on target datasets. Not surprisingly, UPG$^\star$ outperforms UPG and SYN by a wide margin, which implies that fine-tuning the pre-trained model on annotated target datasets is critical for achieving the best possible segmentation performance. The generalization gap of deep learning models across different datasets becomes more evident in Table \ref{table:cross_dataset}, where the highest FBD is only around 83\% if the model fined-tuned on the LSC dataset is tested on our Cannabis dataset. These results re-confirm the difficult of training models that are generalizable across different plant species without fine-tuning on target datasets, which highlights the necessity and importance of developing self-supervised leaf segmentation methods.

As for the self-supervised segmentation methods, we evaluate three most related methods, SSSLIC \cite{kanezaki2018unsupervised} based on the SLIC \cite{achanta2012slic} superpixel algorithm, SSCL \cite{kim2020unsupervised} based on the continuity loss, and our proposed self-supervised segmentation method based on the fully-connected CRF model (SSCRF). We set the superpixel number to 10000 for SSSLIC and the weighting factor of the continuity loss to 5 for SSCL in our experiments. With the semantic segmentation results output by these three methods, we apply the same color-based method to obtain the final leaf segmentation results. We can see from Table \ref{table:natural} that our proposed SSCRF not only consistently outperforms the other two self-supervised methods but also achieves overall better performance than the state-of-the-art unsupervised and supervised methods. Further investigation on the visualization results in Fig. \ref{eqn:greenness} shows that SSSLIC and SSCL tend to merge small leaves into the background, thus leading to the significant performance decline on the subsets `A2' and `A4' of the LSC dataset. The inferior performance of SSSLIC and SSCL can be attributed to the fact that, they only assign the same label to spatially adjacent pixels but lack an effective mechanism to prevent the occurrence of assigning the same label to distinctly different pixels/superpixels. While in our proposed SSCRF, such mechanism is realized via dynamically modeling the pairwise pixel affinities and penalizing inappropriate label assignments to neighboring pixels with large color differences. 
\begin{table}
\centering
\caption{Leaf segmentation results in terms of FBD(\%) on the ``Yellow'' testing sets of the LSC dataset and our Cannabis dataset. `$^\star$' indicates that the pre-trained model is fine-tuned on the \emph{training set} of the LSC dataset.}
\small
\vspace{2pt}
\setlength\tabcolsep{3.5pt}\renewcommand\arraystretch{1.25}
  \noindent\makebox[\textwidth]{%
    \begin{tabular}{lcccccccccc}
      \Xhline{0.3ex}
      \multirow{2}{*}{Dataset}
                   & \multicolumn{3}{c}{Unsupervised} & \multicolumn{4}{c}{Supervised} & \multicolumn{3}{c}{Self-supervised} \\ \cmidrule(lr) {2-4} \cmidrule(lr){5-8} \cmidrule(lr){9-11}
           & EM    & MCS   & Nott. & DC$^\star$   & SYN  & UPG & UPG$^\star$ & SSSLIC  & SSCL    & SSCRF \\  \Xhline{0.2ex}
      Cnbs & 13.8  & 81.6  & 86.8  & -- & 62.3 & 28.2    & --      & 76.3    & 82.1    & \textbf{87.1} \\ \hline
      A1   & 38.7  & 73.6  & 87.8  & 87.4 & 88.8 & 59.3    & \textbf{89.3}    & 87.2    & 85.0    & 88.7 \\ \hline
      A2   & 56.3  & 79.1  & 88.2  & 77.2 & 87.0 & 12.5    & 90.7    & 56.7    & 77.0    & \textbf{92.5} \\ \hline
      A3   & 27.4  & 19.1  & 74.1  & 65.0 & 67.5 & 26.4    & 90.3    & 91.3    & 91.6    & \textbf{93.9} \\ \hline
      A4   & 51.8  & 80.0  & 89.1  & 63.0 & 77.3 & 35.4    & 91.9    & 77.6    & 83.8    & \textbf{92.3} \\ \Xhline{0.3ex}
    \end{tabular}
}%
\label{table:yellow}
\end{table}

\begin{figure*}[]
\captionsetup[subfigure]{labelformat=empty}
         \centering
         \begin{subfigure}{0.2\textwidth}
                 \vspace{-120pt}
                 \caption{}
                 \vspace{-150pt}
         \end{subfigure}%
         \begin{subfigure}{0.2\textwidth}
                 \vspace{-120pt}
                 \caption{Unsupervised}
                 \vspace{-150pt}
         \end{subfigure}%
         \begin{subfigure}{0.3\textwidth}
                 \vspace{-120pt}
                 \caption{Supervised}
                 \vspace{-150pt}
         \end{subfigure}%
         \begin{subfigure}{0.3\textwidth}
                 \vspace{-120pt}
                 \caption{Self-supervised}
                 \vspace{-150pt}
         \end{subfigure}
         \begin{subfigure}{0.1\textwidth}
                 \vspace{-120pt}
                 \caption{}
                 \vspace{-150pt}
         \end{subfigure}%
         \begin{subfigure}{0.2\textwidth}
                 \vspace{-2pt}
                 $\;\;\;\;\;\;\;\;\overbrace{\;\qquad\qquad}$
                 \caption{}
                 \vspace{-30pt}
         \end{subfigure}%
         \begin{subfigure}{0.3\textwidth}
                 \vspace{-2pt}
                 $\;\;\;\;\;\;\;\overbrace{\qquad\qquad\qquad\qquad}$
                 \caption{}
                 \vspace{-30pt}
         \end{subfigure}%
         \begin{subfigure}{0.25\textwidth}
                 \vspace{-2pt}
                 $\;\;\;\;\;\;\;\overbrace{\qquad\qquad\qquad\qquad}$
                 \caption{}
                 \vspace{-30pt}
         \end{subfigure}
        \begin{subfigure}{0.1\textwidth}
                 \vspace{-2pt}
                 \caption{Image}
                 \vspace{-5pt}
         \end{subfigure}%
         \begin{subfigure}{0.1\textwidth}
                 \vspace{-2pt}
                 \caption{Corrected}
                 \vspace{-5pt}
         \end{subfigure}%
         \begin{subfigure}{0.1\textwidth}
                 \vspace{-2pt}
                 \caption{\scriptsize{MCS}}
                 \vspace{-6pt}
         \end{subfigure}%
         \begin{subfigure}{0.1\textwidth}
                 \vspace{-2pt}
                 \caption{\scriptsize{Nott.}}
                 \vspace{-6pt}
         \end{subfigure}%
         \begin{subfigure}{0.1\textwidth}
                 \vspace{-2pt}
                 \caption{DC$^\star$}
                 \vspace{-5pt}
         \end{subfigure}%
         \begin{subfigure}{0.1\textwidth}
                 \vspace{-2pt}
                 \caption{\scriptsize{SYN}}
                 \vspace{-6pt}
         \end{subfigure}%
         \begin{subfigure}{0.1\textwidth}
                 \vspace{-2pt}
                 \caption{\scriptsize{UPG$^\star$}}
                 \vspace{-6pt}
         \end{subfigure}%
         \begin{subfigure}{0.1\textwidth}
                 \vspace{-2pt}
                 \caption{\scriptsize{SSSLIC}}
                 \vspace{-6pt}
         \end{subfigure}%
         \begin{subfigure}{0.1\textwidth}
                 \vspace{-2pt}
                 \caption{\scriptsize{SSCL}}
                 \vspace{-6pt}
         \end{subfigure}%
         \begin{subfigure}{0.1\textwidth}
                 \vspace{-2pt}
                 \caption{\scriptsize{SSCRF}}
                 \vspace{-6pt}
         \end{subfigure}
         \vspace{1pt}
         \begin{subfigure}{0.1\textwidth}
                 \includegraphics[width=0.98\linewidth]{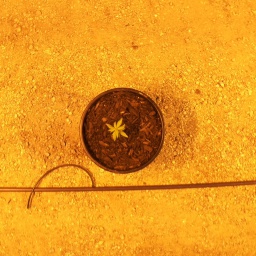}
         \end{subfigure}%
         \begin{subfigure}{0.1\textwidth}
                 \includegraphics[width=0.98\linewidth]{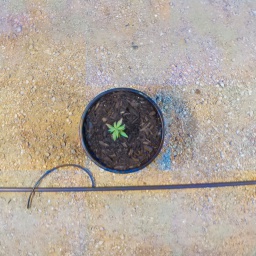}
         \end{subfigure}%
         \begin{subfigure}{0.1\textwidth}
                 \includegraphics[width=0.98\linewidth]{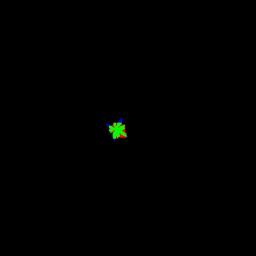}
         \end{subfigure}%
         \begin{subfigure}{0.1\textwidth}
                 \includegraphics[width=0.98\linewidth]{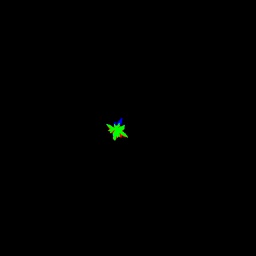}
         \end{subfigure}%
         \begin{subfigure}{0.1\textwidth}
                 \includegraphics[width=0.98\linewidth]{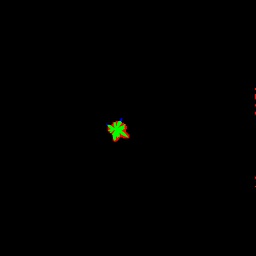}
         \end{subfigure}%
         \begin{subfigure}{0.1\textwidth}
                 \includegraphics[width=0.98\linewidth]{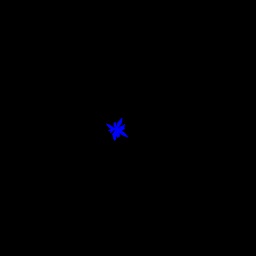}
         \end{subfigure}
         \begin{subfigure}{0.1\textwidth}
                 \includegraphics[width=0.98\linewidth]{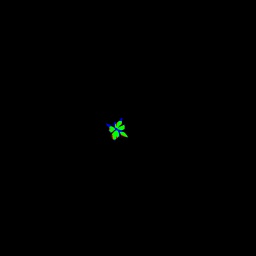}
         \end{subfigure}%
         \begin{subfigure}{0.1\textwidth}
                 \includegraphics[width=0.98\linewidth]{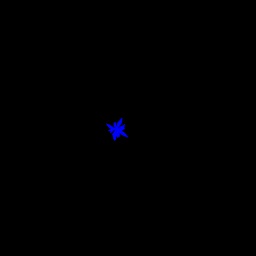}
         \end{subfigure}%
         \begin{subfigure}{0.1\textwidth}
                 \includegraphics[width=0.98\linewidth]{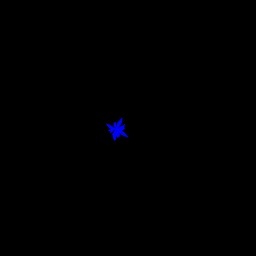}
         \end{subfigure}%
         \begin{subfigure}{0.1\textwidth}
                 \includegraphics[width=0.98\linewidth]{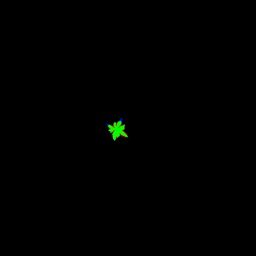}
         \end{subfigure}
         \vspace{1pt}
         \begin{subfigure}{0.1\textwidth}
                 \includegraphics[width=0.98\linewidth]{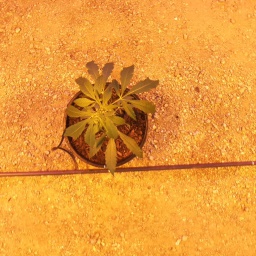}
         \end{subfigure}%
         \begin{subfigure}{0.1\textwidth}
                 \includegraphics[width=0.98\linewidth]{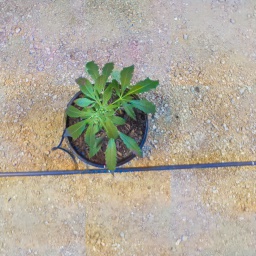}
         \end{subfigure}%
         \begin{subfigure}{0.1\textwidth}
                 \includegraphics[width=0.98\linewidth]{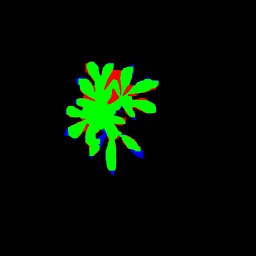}
         \end{subfigure}%
         \begin{subfigure}{0.1\textwidth}
                 \includegraphics[width=0.98\linewidth]{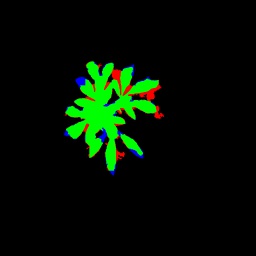}
         \end{subfigure}%
         \begin{subfigure}{0.1\textwidth}
                 \includegraphics[width=0.98\linewidth]{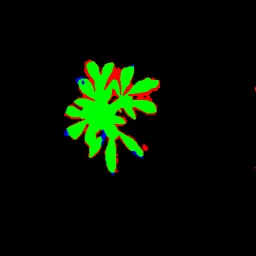}
         \end{subfigure}%
         \begin{subfigure}{0.1\textwidth}
                 \includegraphics[width=0.98\linewidth]{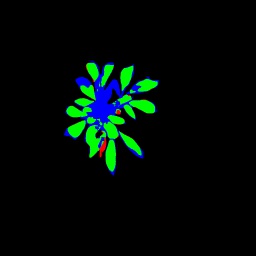}
         \end{subfigure}
         \begin{subfigure}{0.1\textwidth}
                 \includegraphics[width=0.98\linewidth]{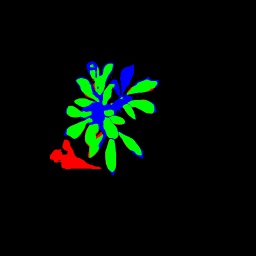}
         \end{subfigure}%
         \begin{subfigure}{0.1\textwidth}
                 \includegraphics[width=0.98\linewidth]{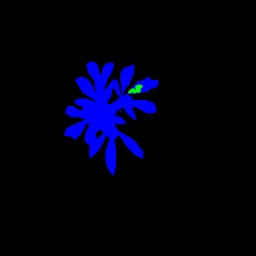}
         \end{subfigure}%
         \begin{subfigure}{0.1\textwidth}
                 \includegraphics[width=0.98\linewidth]{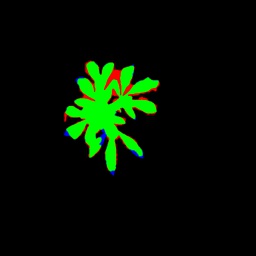}
         \end{subfigure}%
         \begin{subfigure}{0.1\textwidth}
                 \includegraphics[width=0.98\linewidth]{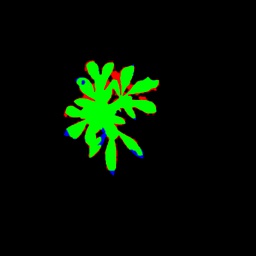}
         \end{subfigure}
         \vspace{1pt}
         \begin{subfigure}{0.1\textwidth}
                 \includegraphics[width=0.98\linewidth]{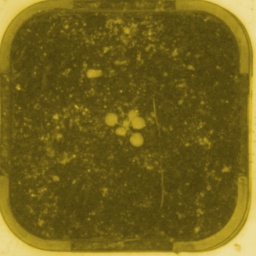}
         \end{subfigure}%
         \begin{subfigure}{0.1\textwidth}
                 \includegraphics[width=0.98\linewidth]{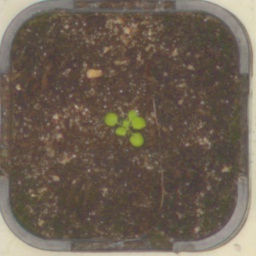}
         \end{subfigure}%
         \begin{subfigure}{0.1\textwidth}
                 \includegraphics[width=0.98\linewidth]{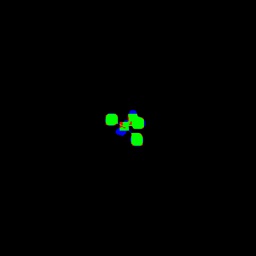}
         \end{subfigure}%
         \begin{subfigure}{0.1\textwidth}
                 \includegraphics[width=0.98\linewidth]{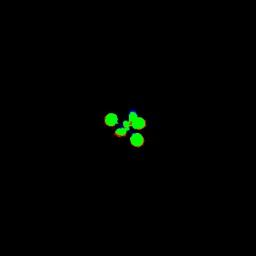}
         \end{subfigure}%
         \begin{subfigure}{0.1\textwidth}
                 \includegraphics[width=0.98\linewidth]{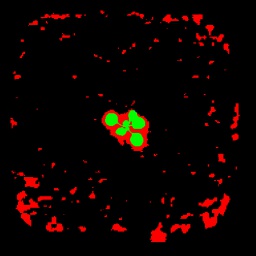}
         \end{subfigure}%
         \begin{subfigure}{0.1\textwidth}
                 \includegraphics[width=0.98\linewidth]{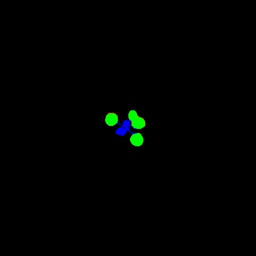}
         \end{subfigure}%
         \begin{subfigure}{0.1\textwidth}
                 \includegraphics[width=0.98\linewidth]{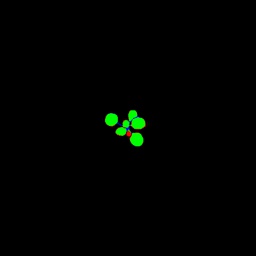}
         \end{subfigure}%
         \begin{subfigure}{0.1\textwidth}
                 \includegraphics[width=0.98\linewidth]{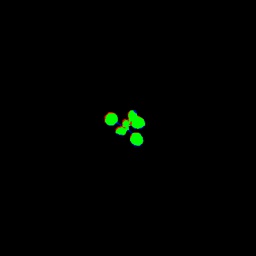}
         \end{subfigure}%
         \begin{subfigure}{0.1\textwidth}
                 \includegraphics[width=0.98\linewidth]{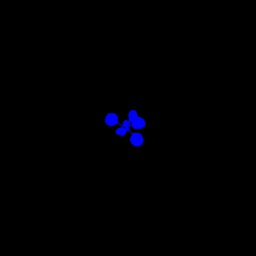}
         \end{subfigure}%
         \begin{subfigure}{0.1\textwidth}
                 \includegraphics[width=0.98\linewidth]{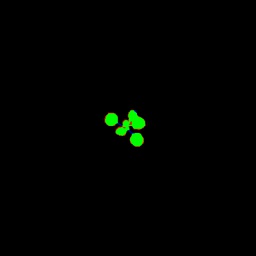}
         \end{subfigure}
         \vspace{1pt}
         \begin{subfigure}{0.1\textwidth}
                 \includegraphics[width=0.98\linewidth]{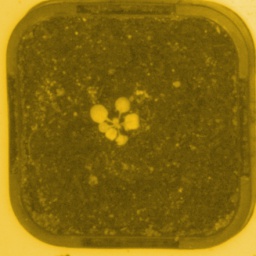}
         \end{subfigure}%
         \begin{subfigure}{0.1\textwidth}
                 \includegraphics[width=0.98\linewidth]{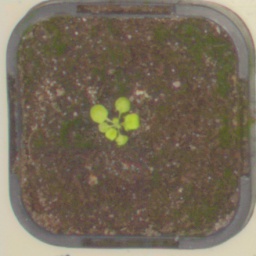}
         \end{subfigure}%
         \begin{subfigure}{0.1\textwidth}
                 \includegraphics[width=0.98\linewidth]{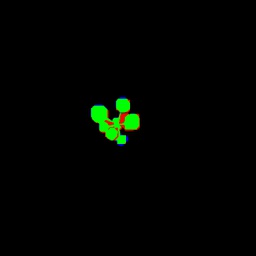}
         \end{subfigure}%
         \begin{subfigure}{0.1\textwidth}
                 \includegraphics[width=0.98\linewidth]{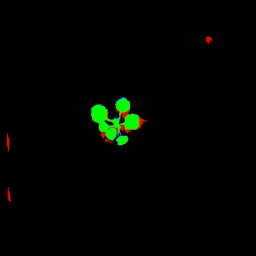}
         \end{subfigure}%
         \begin{subfigure}{0.1\textwidth}
                 \includegraphics[width=0.98\linewidth]{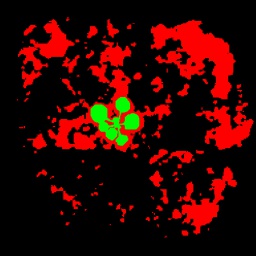}
         \end{subfigure}%
         \begin{subfigure}{0.1\textwidth}
                 \includegraphics[width=0.98\linewidth]{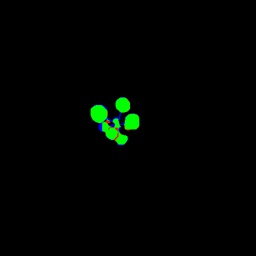}
         \end{subfigure}%
         \begin{subfigure}{0.1\textwidth}
                 \includegraphics[width=0.98\linewidth]{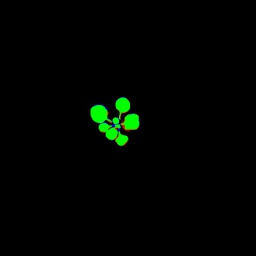}
         \end{subfigure}%
         \begin{subfigure}{0.1\textwidth}
                 \includegraphics[width=0.98\linewidth]{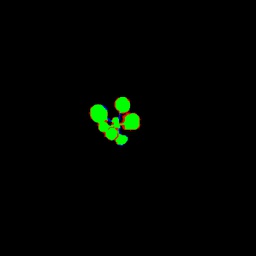}
         \end{subfigure}%
         \begin{subfigure}{0.1\textwidth}
                 \includegraphics[width=0.98\linewidth]{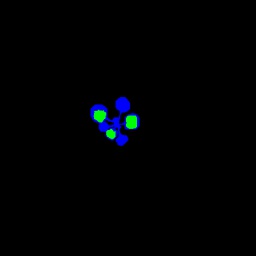}
         \end{subfigure}%
         \begin{subfigure}{0.1\textwidth}
                 \includegraphics[width=0.98\linewidth]{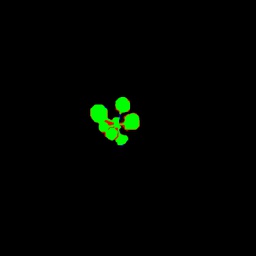}
         \end{subfigure}
         \vspace{1pt}
         \begin{subfigure}{0.1\textwidth}
                 \includegraphics[width=0.98\linewidth]{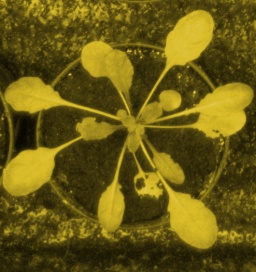}
         \end{subfigure}%
         \begin{subfigure}{0.1\textwidth}
                 \includegraphics[width=0.98\linewidth]{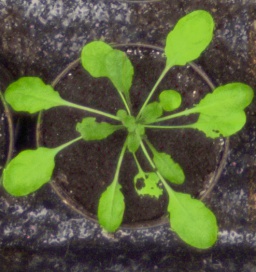}
         \end{subfigure}%
         \begin{subfigure}{0.1\textwidth}
                 \includegraphics[width=0.98\linewidth]{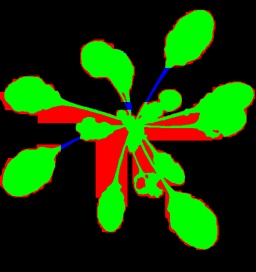}
         \end{subfigure}%
         \begin{subfigure}{0.1\textwidth}
                 \includegraphics[width=0.98\linewidth]{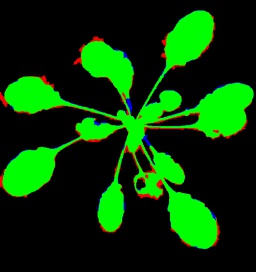}
         \end{subfigure}%
         \begin{subfigure}{0.1\textwidth}
                 \includegraphics[width=0.98\linewidth]{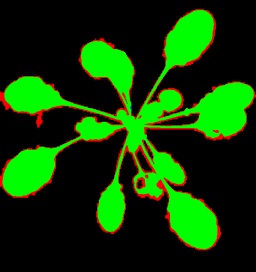}
         \end{subfigure}%
         \begin{subfigure}{0.1\textwidth}
                 \includegraphics[width=0.98\linewidth]{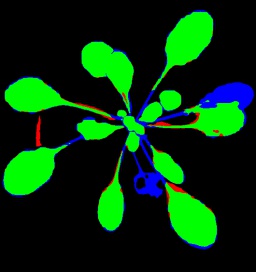}
         \end{subfigure}%
         \begin{subfigure}{0.1\textwidth}
                 \includegraphics[width=0.98\linewidth]{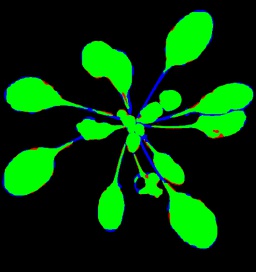}
         \end{subfigure}%
         \begin{subfigure}{0.1\textwidth}
                 \includegraphics[width=0.98\linewidth]{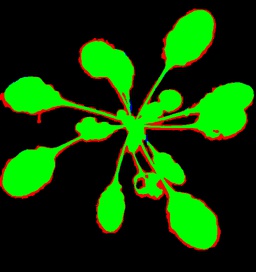}
         \end{subfigure}%
         \begin{subfigure}{0.1\textwidth}
                 \includegraphics[width=0.98\linewidth]{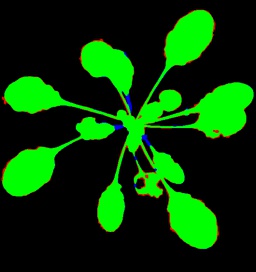}
         \end{subfigure}%
         \begin{subfigure}{0.1\textwidth}
                 \includegraphics[width=0.98\linewidth]{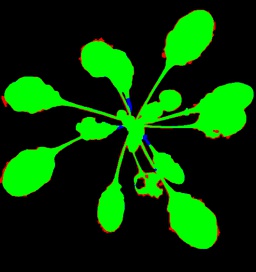}
         \end{subfigure}
         \vspace{1pt}
         \begin{subfigure}{0.1\textwidth}
                 \includegraphics[width=0.98\linewidth]{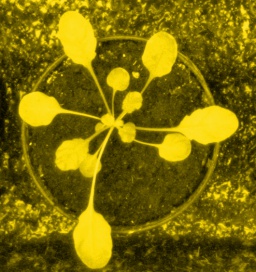}
         \end{subfigure}%
         \begin{subfigure}{0.1\textwidth}
                 \includegraphics[width=0.98\linewidth]{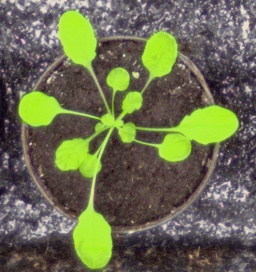}
         \end{subfigure}%
         \begin{subfigure}{0.1\textwidth}
                 \includegraphics[width=0.98\linewidth]{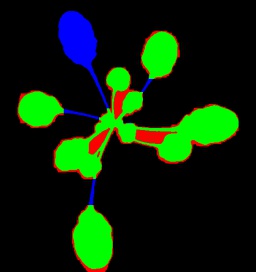}
         \end{subfigure}%
         \begin{subfigure}{0.1\textwidth}
                 \includegraphics[width=0.98\linewidth]{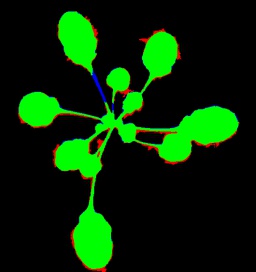}
         \end{subfigure}%
         \begin{subfigure}{0.1\textwidth}
                 \includegraphics[width=0.98\linewidth]{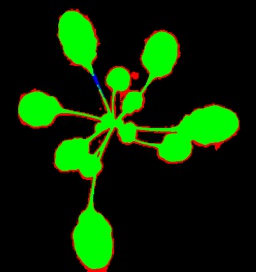}
         \end{subfigure}%
         \begin{subfigure}{0.1\textwidth}
                 \includegraphics[width=0.98\linewidth]{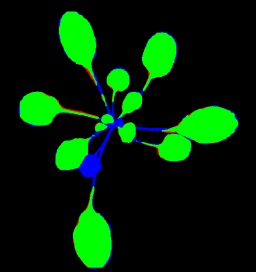}
         \end{subfigure}%
         \begin{subfigure}{0.1\textwidth}
                 \includegraphics[width=0.98\linewidth]{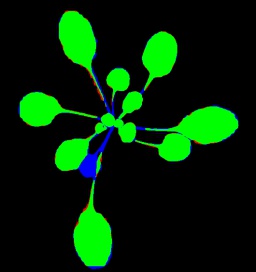}
         \end{subfigure}%
         \begin{subfigure}{0.1\textwidth}
                 \includegraphics[width=0.98\linewidth]{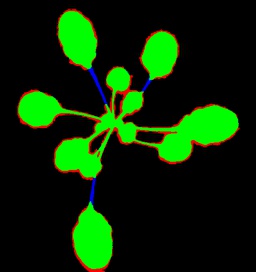}
         \end{subfigure}%
         \begin{subfigure}{0.1\textwidth}
                 \includegraphics[width=0.98\linewidth]{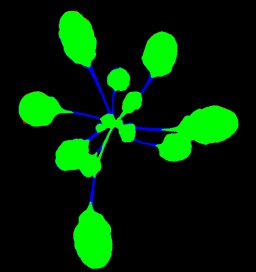}
         \end{subfigure}%
         \begin{subfigure}{0.1\textwidth}
                 \includegraphics[width=0.98\linewidth]{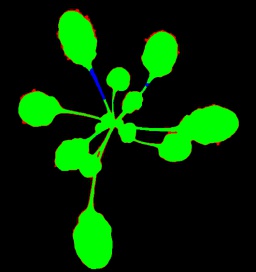}
         \end{subfigure}
         \vspace{1pt}
         \begin{subfigure}{0.1\textwidth}
                 \includegraphics[width=0.98\linewidth]{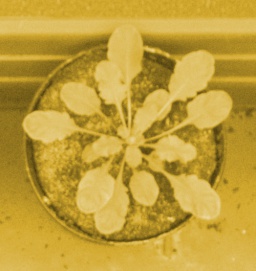}
         \end{subfigure}%
         \begin{subfigure}{0.1\textwidth}
                 \includegraphics[width=0.98\linewidth]{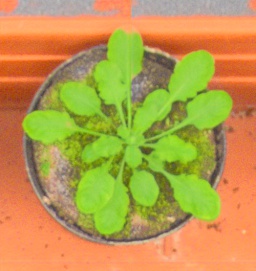}
         \end{subfigure}%
         \begin{subfigure}{0.1\textwidth}
                 \includegraphics[width=0.98\linewidth]{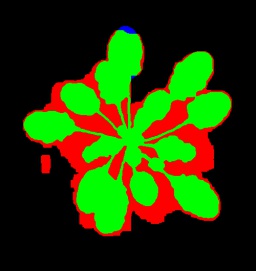}
         \end{subfigure}%
         \begin{subfigure}{0.1\textwidth}
                 \includegraphics[width=0.98\linewidth]{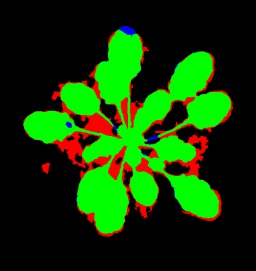}
         \end{subfigure}%
         \begin{subfigure}{0.1\textwidth}
                 \includegraphics[width=0.98\linewidth]{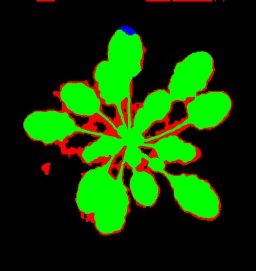}
         \end{subfigure}%
         \begin{subfigure}{0.1\textwidth}
                 \includegraphics[width=0.98\linewidth]{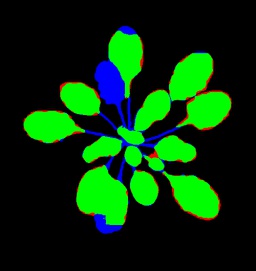}
         \end{subfigure}%
         \begin{subfigure}{0.1\textwidth}
                 \includegraphics[width=0.98\linewidth]{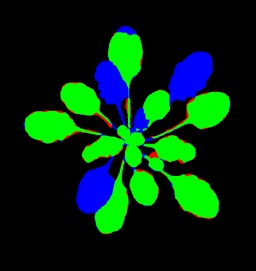}
         \end{subfigure}%
         \begin{subfigure}{0.1\textwidth}
                 \includegraphics[width=0.98\linewidth]{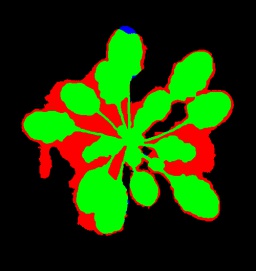}
         \end{subfigure}%
         \begin{subfigure}{0.1\textwidth}
                 \includegraphics[width=0.98\linewidth]{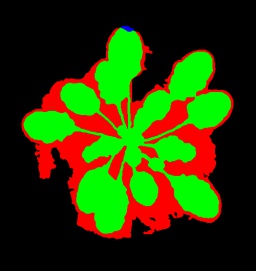}
         \end{subfigure}%
         \begin{subfigure}{0.1\textwidth}
                 \includegraphics[width=0.98\linewidth]{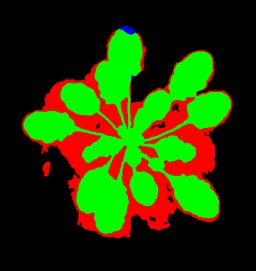}
         \end{subfigure}
         \vspace{1pt}
         \begin{subfigure}{0.1\textwidth}
                 \includegraphics[width=0.98\linewidth]{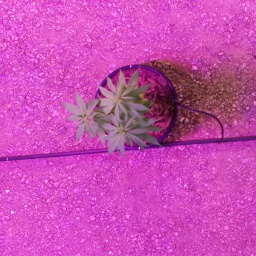}
         \end{subfigure}%
         \begin{subfigure}{0.1\textwidth}
                 \includegraphics[width=0.98\linewidth]{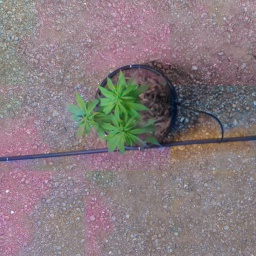}
         \end{subfigure}%
         \begin{subfigure}{0.1\textwidth}
                 \includegraphics[width=0.98\linewidth]{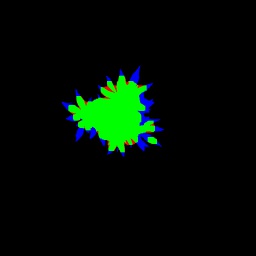}
         \end{subfigure}%
         \begin{subfigure}{0.1\textwidth}
                 \includegraphics[width=0.98\linewidth]{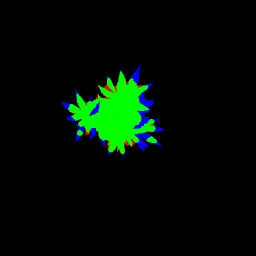}
         \end{subfigure}%
         \begin{subfigure}{0.1\textwidth}
                 \includegraphics[width=0.98\linewidth]{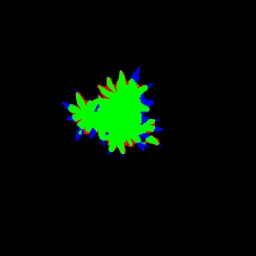}
         \end{subfigure}%
         \begin{subfigure}{0.1\textwidth}
                 \includegraphics[width=0.98\linewidth]{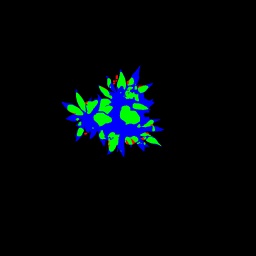}
         \end{subfigure}
         \begin{subfigure}{0.1\textwidth}
                 \includegraphics[width=0.98\linewidth]{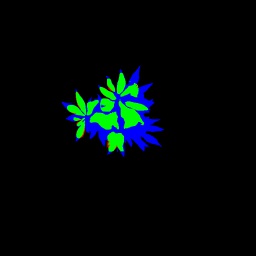}
         \end{subfigure}%
         \begin{subfigure}{0.1\textwidth}
                 \includegraphics[width=0.98\linewidth]{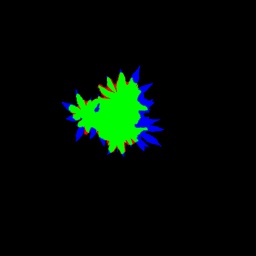}
         \end{subfigure}%
         \begin{subfigure}{0.1\textwidth}
                 \includegraphics[width=0.98\linewidth]{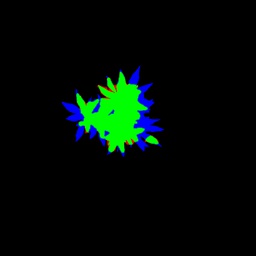}
         \end{subfigure}%
         \begin{subfigure}{0.1\textwidth}
                 \includegraphics[width=0.98\linewidth]{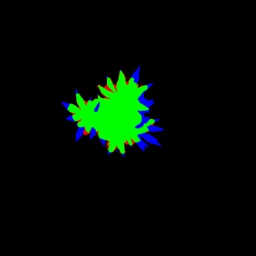}
         \end{subfigure}
         \vspace{1pt}
         \begin{subfigure}{0.1\textwidth}
                 \includegraphics[width=0.98\linewidth]{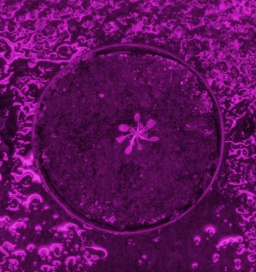}
         \end{subfigure}%
         \begin{subfigure}{0.1\textwidth}
                 \includegraphics[width=0.98\linewidth]{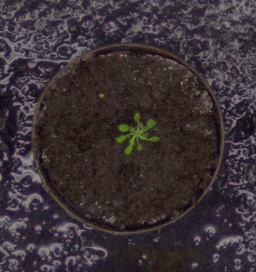}
         \end{subfigure}%
         \begin{subfigure}{0.1\textwidth}
                 \includegraphics[width=0.98\linewidth]{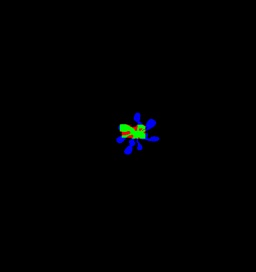}
         \end{subfigure}%
         \begin{subfigure}{0.1\textwidth}
                 \includegraphics[width=0.98\linewidth]{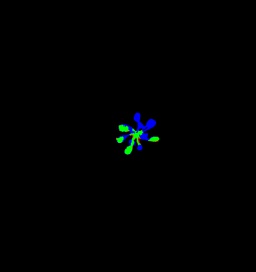}
         \end{subfigure}%
         \begin{subfigure}{0.1\textwidth}
                 \includegraphics[width=0.98\linewidth]{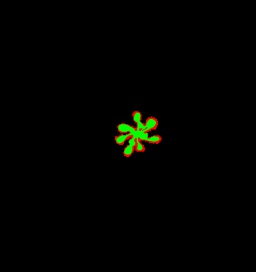}
         \end{subfigure}%
         \begin{subfigure}{0.1\textwidth}
                 \includegraphics[width=0.98\linewidth]{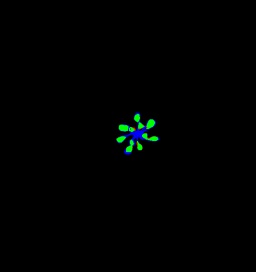}
         \end{subfigure}%
         \begin{subfigure}{0.1\textwidth}
                 \includegraphics[width=0.98\linewidth]{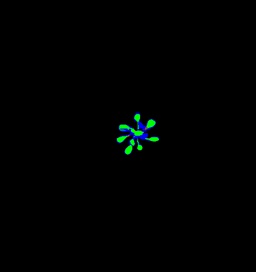}
         \end{subfigure}%
         \begin{subfigure}{0.1\textwidth}
                 \includegraphics[width=0.98\linewidth]{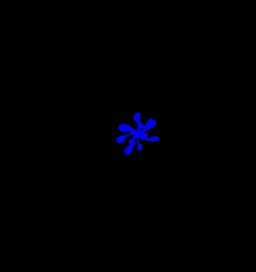}
         \end{subfigure}%
         \begin{subfigure}{0.1\textwidth}
                 \includegraphics[width=0.98\linewidth]{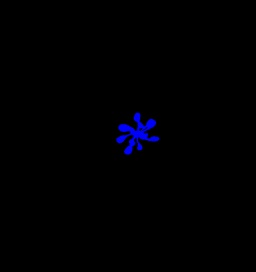}
         \end{subfigure}%
         \begin{subfigure}{0.1\textwidth}
                 \includegraphics[width=0.98\linewidth]{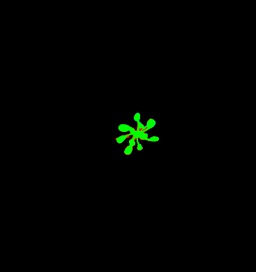}
         \end{subfigure}
         \vspace{1pt}
         \begin{subfigure}{0.1\textwidth}
                 \includegraphics[width=0.98\linewidth]{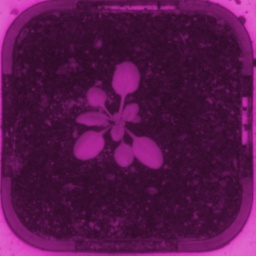}
         \end{subfigure}%
         \begin{subfigure}{0.1\textwidth}
                 \includegraphics[width=0.98\linewidth]{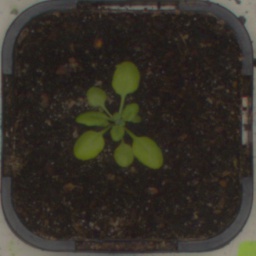}
         \end{subfigure}%
         \begin{subfigure}{0.1\textwidth}
                 \includegraphics[width=0.98\linewidth]{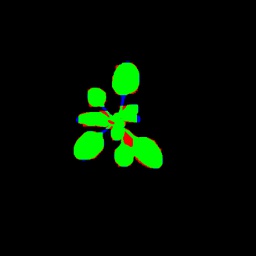}
         \end{subfigure}%
         \begin{subfigure}{0.1\textwidth}
                 \includegraphics[width=0.98\linewidth]{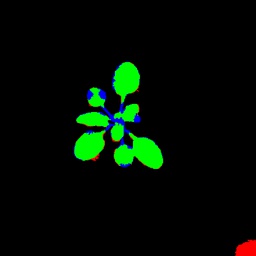}
         \end{subfigure}%
         \begin{subfigure}{0.1\textwidth}
                 \includegraphics[width=0.98\linewidth]{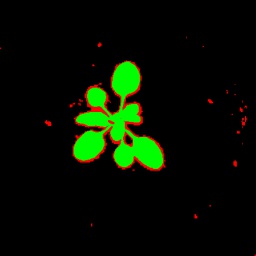}
         \end{subfigure}%
         \begin{subfigure}{0.1\textwidth}
                 \includegraphics[width=0.98\linewidth]{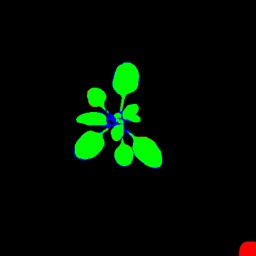}
         \end{subfigure}%
         \begin{subfigure}{0.1\textwidth}
                 \includegraphics[width=0.98\linewidth]{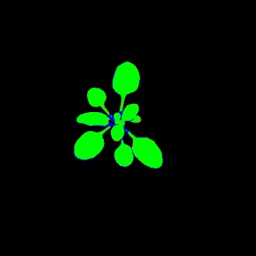}
         \end{subfigure}%
         \begin{subfigure}{0.1\textwidth}
                 \includegraphics[width=0.98\linewidth]{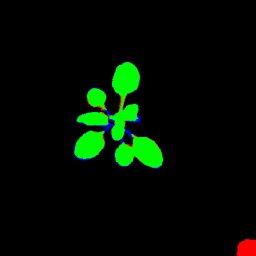}
         \end{subfigure}%
         \begin{subfigure}{0.1\textwidth}
                 \includegraphics[width=0.98\linewidth]{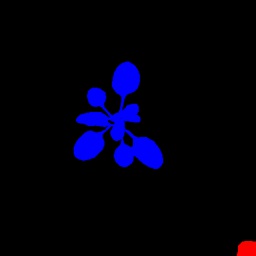}
         \end{subfigure}%
         \begin{subfigure}{0.1\textwidth}
                 \includegraphics[width=0.98\linewidth]{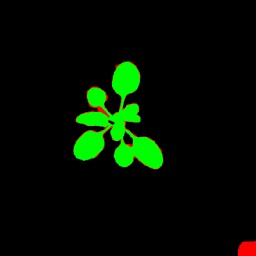}
         \end{subfigure}
         \vspace{1pt}
         \begin{subfigure}{0.1\textwidth}
                 \includegraphics[width=0.98\linewidth]{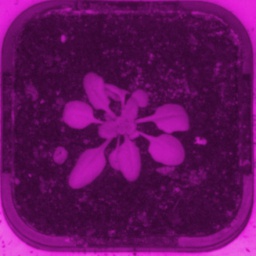}
         \end{subfigure}%
         \begin{subfigure}{0.1\textwidth}
                 \includegraphics[width=0.98\linewidth]{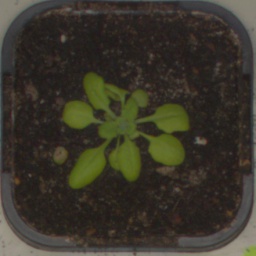}
         \end{subfigure}%
         \begin{subfigure}{0.1\textwidth}
                 \includegraphics[width=0.98\linewidth]{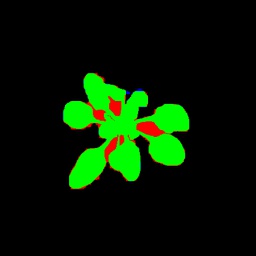}
         \end{subfigure}%
         \begin{subfigure}{0.1\textwidth}
                 \includegraphics[width=0.98\linewidth]{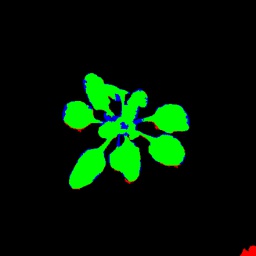}
         \end{subfigure}%
         \begin{subfigure}{0.1\textwidth}
                 \includegraphics[width=0.98\linewidth]{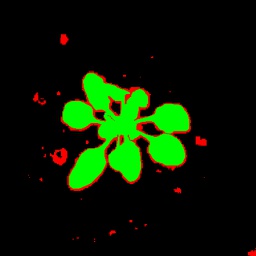}
         \end{subfigure}%
         \begin{subfigure}{0.1\textwidth}
                 \includegraphics[width=0.98\linewidth]{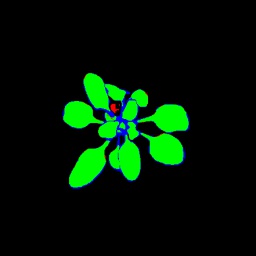}
         \end{subfigure}%
         \begin{subfigure}{0.1\textwidth}
                 \includegraphics[width=0.98\linewidth]{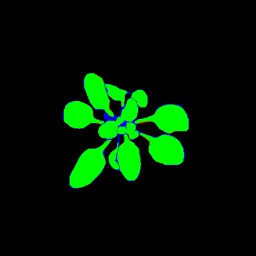}
         \end{subfigure}%
         \begin{subfigure}{0.1\textwidth}
                 \includegraphics[width=0.98\linewidth]{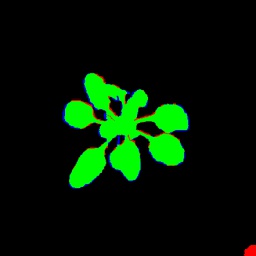}
         \end{subfigure}%
         \begin{subfigure}{0.1\textwidth}
                 \includegraphics[width=0.98\linewidth]{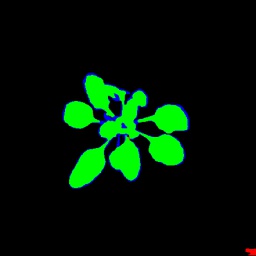}
         \end{subfigure}%
         \begin{subfigure}{0.1\textwidth}
                 \includegraphics[width=0.98\linewidth]{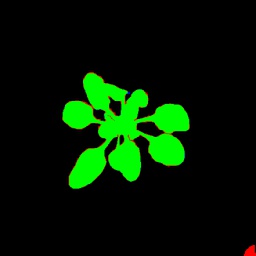}
         \end{subfigure}
         \vspace{1pt}
         \begin{subfigure}{0.1\textwidth}
                 \includegraphics[width=0.98\linewidth]{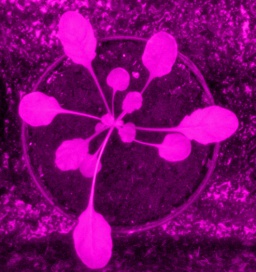}
         \end{subfigure}%
         \begin{subfigure}{0.1\textwidth}
                 \includegraphics[width=0.98\linewidth]{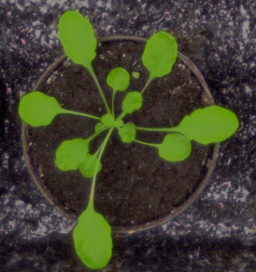}
         \end{subfigure}%
         \begin{subfigure}{0.1\textwidth}
                 \includegraphics[width=0.98\linewidth]{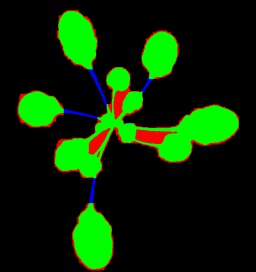}
         \end{subfigure}%
         \begin{subfigure}{0.1\textwidth}
                 \includegraphics[width=0.98\linewidth]{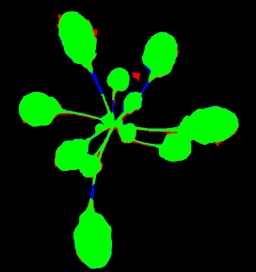}
         \end{subfigure}%
         \begin{subfigure}{0.1\textwidth}
                 \includegraphics[width=0.98\linewidth]{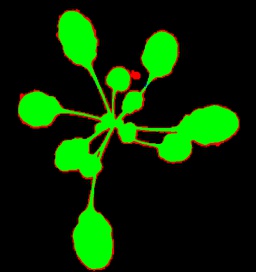}
         \end{subfigure}%
         \begin{subfigure}{0.1\textwidth}
                 \includegraphics[width=0.98\linewidth]{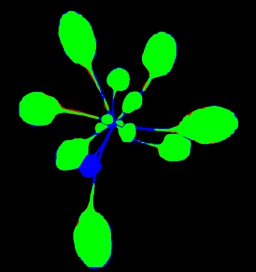}
         \end{subfigure}%
         \begin{subfigure}{0.1\textwidth}
                 \includegraphics[width=0.98\linewidth]{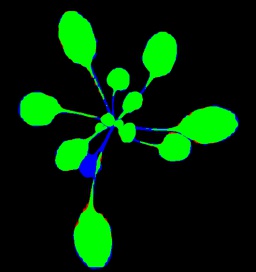}
         \end{subfigure}%
         \begin{subfigure}{0.1\textwidth}
                 \includegraphics[width=0.98\linewidth]{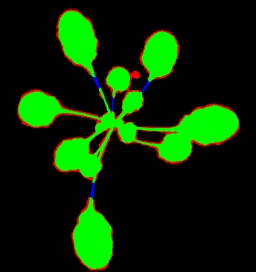}
         \end{subfigure}%
         \begin{subfigure}{0.1\textwidth}
                 \includegraphics[width=0.98\linewidth]{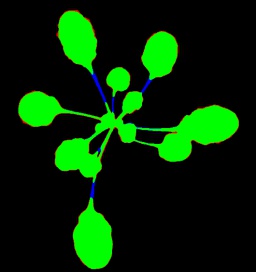}
         \end{subfigure}%
         \begin{subfigure}{0.1\textwidth}
                 \includegraphics[width=0.98\linewidth]{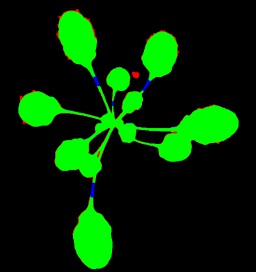}
         \end{subfigure}
         \vspace{1pt}
         \begin{subfigure}{0.1\textwidth}
                 \includegraphics[width=0.98\linewidth]{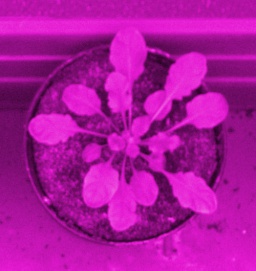}
         \end{subfigure}%
         \begin{subfigure}{0.1\textwidth}
                 \includegraphics[width=0.98\linewidth]{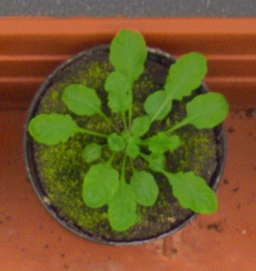}
         \end{subfigure}%
         \begin{subfigure}{0.1\textwidth}
                 \includegraphics[width=0.98\linewidth]{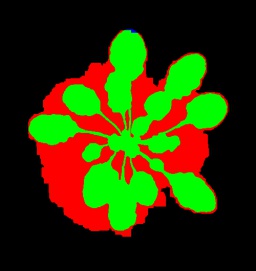}
         \end{subfigure}%
         \begin{subfigure}{0.1\textwidth}
                 \includegraphics[width=0.98\linewidth]{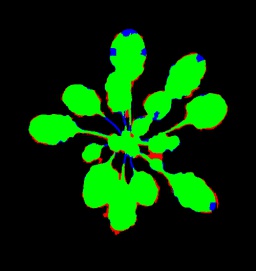}
         \end{subfigure}%
         \begin{subfigure}{0.1\textwidth}
                 \includegraphics[width=0.98\linewidth]{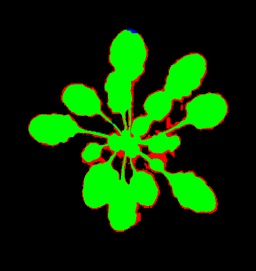}
         \end{subfigure}%
         \begin{subfigure}{0.1\textwidth}
                 \includegraphics[width=0.98\linewidth]{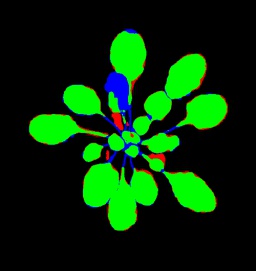}
         \end{subfigure}%
         \begin{subfigure}{0.1\textwidth}
                 \includegraphics[width=0.98\linewidth]{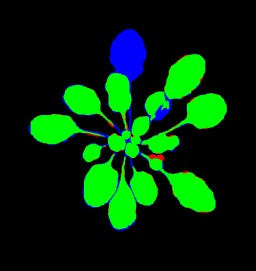}
         \end{subfigure}%
         \begin{subfigure}{0.1\textwidth}
                 \includegraphics[width=0.98\linewidth]{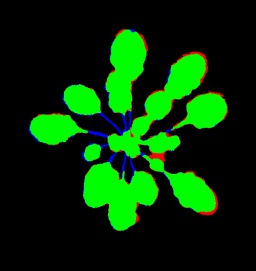}
         \end{subfigure}%
         \begin{subfigure}{0.1\textwidth}
                 \includegraphics[width=0.98\linewidth]{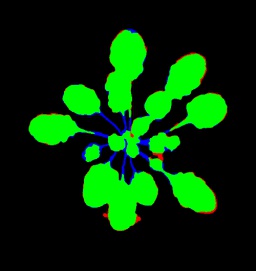}
         \end{subfigure}%
         \begin{subfigure}{0.1\textwidth}
                 \includegraphics[width=0.98\linewidth]{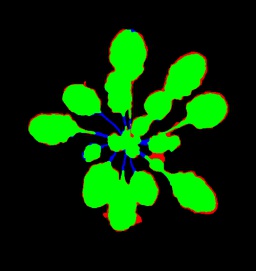}
         \end{subfigure}
         \caption{Example leaf segmentation results on ``Yellow'' and ``Purple'' images. Color coding: \textcolor{OliveGreen}{\emph{green}}: detected leaf regions (true positives); \textcolor{red}{\emph{red}}: detected non-leaf regions (false positives); \textcolor{blue}{\emph{blue}}: mis-detected leaf regions (false negatives).}
\label{fig:leaf_segmentation_examples_cc}
\vspace{-1pt}
\end{figure*}

To evaluate the performance of the proposed color correction model, we apply color correction to the images in the ``Yellow'' and ``Purple'' \emph{testing sets} of the LSC dataset and our Cannabis dataset with the color correction models trained on the corresponding ``Natural'' \emph{training sets}. We then repeat the above leaf segmentation experiments for all compared methods on the color-corrected images of the ``Yellow'' and ``Purple'' \emph{testing sets}. The quantitative results are reported in Table \ref{table:yellow} and Table \ref{table:purple}, while some qualitative results can be found in Fig. \ref{fig:leaf_segmentation_examples_cc}, where we also show the color-corrected images in the second column to visualize the color correction performance. We exclude EM \cite{leafsnap} and UPG \cite{UPGen} in  Fig. \ref{fig:leaf_segmentation_examples_cc} because of their poor performance. We would like to make a few remarks for these results: 1) While trained on the same ``Natural'' \emph{training sets}, the color correction model exhibits somewhat performance variations on images taken under different lighting conditions. It is generally easier to correct color for ``Yellow'' images than ``Purple'' images, probably because the yellow color is statistically distributed closer to the green color in the $L^*a^*b^*$ color space. 2) For the leaf segmentation task, it is important to include in the training set the images collected at various growth stages covering the whole life cycle of plants. In the training set of our Cannabis dataset, the majority of the images were collected at early growth stages including 72 images that only contain empty plant pots, which, to some extent, compromise the color correction performance on our Cannabis dataset. 
3) Our proposed self-supervised leaf segmentation method still achieves overall better performance than other methods on the color-corrected images across different datasets, which highlights the potential of our method in achieving effective and generalizable leaf segmentation.

\begin{table}
\centering
\caption{Leaf segmentation results in terms of FBD(\%) on the ``Purple'' testing sets of the LSC dataset and our Cannabis dataset. `$^\star$' indicates that the pre-trained model is fine-tuned on the \emph{training set} of the LSC dataset.}
\small
\vspace{2pt}
\setlength\tabcolsep{3.5pt}\renewcommand\arraystretch{1.25}
  \noindent\makebox[\textwidth]{%
    \begin{tabular}{lcccccccccc}
      \Xhline{0.3ex}
      \multirow{2}{*}{Dataset}
                   & \multicolumn{3}{c}{Unsupervised} & \multicolumn{4}{c}{Supervised} & \multicolumn{3}{c}{Self-supervised} \\ \cmidrule(lr) {2-4} \cmidrule(lr){5-8} \cmidrule(lr){9-11}
           & EM    & MCS   & Nott. & DC$^\star$   & SYN  & UPG     & UPG$^\star$ & SSSLIC  & SSCL   & SSCRF \\  \Xhline{0.2ex}
      Cnbs & 28.2  & 82.2  & \textbf{84.8}  & -- & 51.3 & 32.4    & --          & 75.7    & 80.7   & 83.9\\ \hline
      A1   & 37.3  & 72.9  & 87.8  & 91.7 & 90.0 & 48.5    & 89.7        & 91.4    & 92.5   & \textbf{94.7} \\ \hline
      A2   & 61.3  & 77.4  & 88.2  & 80.7 & 77.2 & 24.9    & 88.7        & 63.2    & 69.6   & \textbf{92.6} \\ \hline
      A3   & 22.9  & 18.7  & 81.6  & 68.2 & 67.1 & 45.4    & 89.9        & 87.3    & 91.9   & \textbf{94.8} \\ \hline
      A4   & 56.3  & 74.4  & 82.6  & 75.5 & 77.4 & 28.7    & \textbf{88.4} & 71.3    & 74.8   & 83.8 \\ \Xhline{0.3ex}
    \end{tabular}
}%
\label{table:purple}
\end{table}

\section{Conclusion and Future Work}
\label{sec:conclusion}
In this work, we presented a self-supervised leaf segmentation framework that is capable of segmenting leaf regions from the background under complex illumination conditions without annotated training data. Comprehensive experiments on the CVPPP LSC dataset and our Cannabis dataset demonstrated that the proposed method achieves state-of-the-art performance. Despite its effectiveness in segmenting leaf regions under varying lighting conditions and the generalizability across different plant species, there is still some room for improvement. In the proposed self-supervised semantic segmentation model, the pixel-wise label assignment is updated and refined in an iterative manner, which may require hundreds of iterations to obtain sensible results. For an iteration number of $T{=}300$, the segmentation takes $20{\text -}30$ seconds for an image of size $512{\times}512$ pixels on our desktop PC with a Nvidia GTX 2080Ti 11GB GPU. While this meets the requirements of our plant growth monitoring project, we will seek to improve the efficiency of our method by using heuristic early stopping criteria and initializing the pixel-level embeddings learning network with pre-trained weights. Given the promising results of our self-supervised semantic segmentation, another interesting line of research for future work is to explore the possibility of self-supervised instance-level leaf segmentation by bridging the gap between semantic segmentation and instance segmentation. 

{\small
\bibliography{manuscript}}

\end{document}